\pdfoutput=1
\documentclass[10pt,twocolumn,letterpaper]{article}

\usepackage{wacv}
\usepackage{times}
\usepackage{epsfig}
\usepackage{graphicx}
\usepackage{amsmath}
\usepackage{amssymb}

\usepackage{array}
\usepackage{nicefrac}
\usepackage{algorithm}
\usepackage{algpseudocode}
\usepackage{subfigure}
\usepackage{epstopdf}
\usepackage{multirow}
\usepackage{setspace}
\usepackage{wrapfig}
\usepackage[mathcal]{eucal}
\usepackage[export]{adjustbox}
\usepackage[leftcaption]{sidecap}
\usepackage{scrextend}
\usepackage[table]{xcolor}

\newcommand{\pixel}[1]{\boldsymbol{\mathsf{#1}}}
\newcommand{\opvec}[1]{\mathop{\mathsf{vec}}\left({#1}\right)}
\newcommand{\opvecinv}[1]{\mathop{\mathsf{vec}^{-1}}\left({#1}\right)}

\newcommand{\hist}[1]{\mathop{\mathsf{hist}}\left({#1}\right)}
\newcommand{\dist}[1]{\mathop{\mathsf{dist}}\left({#1}\right)}
\newcommand{\ecc}[1]{\mathop{\mathsf{ecc}}\left({#1}\right)}
\newcommand{\diam}[1]{\mathop{\mathsf{diam}}\left({#1}\right)}
\newcommand{\sv}{\mathbf{w}}
\newcommand{\slack}{\xi}

\newcommand{\vect}[1]{\textbf{#1}}
\newcommand{\tr}{^\intercal}
\newcommand{\argmax}{\mathop{\mathrm{argmax}}}

\def\NoNumber#1{{\def\alglinenumber##1{}\State #1}\addtocounter{ALG@line}{-1}}

\wacvfinalcopy 

\def\wacvPaperID{253} 

\ifwacvfinal\fi
\begin{document}

\title{Progressive Tree-like Curvilinear Structure Reconstruction with Structured Ranking Learning and Graph Algorithm}

\author{Seong-Gyun Jeong \hspace{1cm} Yuliya Tarabalka \hspace{1cm} Nicolas Nisse \hspace{1cm} Josiane Zerubia\\
Universit\'e C\^ote d'Azur, Inria, Sophia Antipolis, France\\
{\tt\small \{firstname.lastname\}@inria.fr}
}

\maketitle
\ifwacvfinal\thispagestyle{empty}\fi

\begin{abstract}
   We propose a novel tree-like curvilinear structure reconstruction algorithm based on supervised learning and graph theory. In this work we analyze image patches to obtain the local major orientations and the rankings that correspond to the curvilinear structure. To extract local curvilinear features, we compute oriented gradient information using steerable filters. We then employ Structured Support Vector Machine for ordinal regression of the input image patches, where the ordering is determined by shape similarity to latent curvilinear structure. Finally, we progressively reconstruct the curvilinear structure by looking for geodesic paths connecting remote vertices in the graph built on the structured output rankings. Experimental results show that the proposed algorithm faithfully provides topological features of the curvilinear structures using minimal pixels for various datasets.
\end{abstract}

\section{Introduction}
\label{sec:introduction}
	Many computer vision algorithms have been proposed to analyze tree-like curvilinear structures (also called line networks), such as blood vessels for retinal images~\cite{Frangi1998,Staal2004}, filament structure of biological images~\cite{Peng2011,Wang2011}, road network for remote sensing~\cite{Hu2007,Lacoste2005,Valero2010}, and defects on materials~\cite{Chambon2010,Iyer2005}. Although human can intuitively perceive the curvilinear structures in very different images and application scenarios, most curvilinear structure detection methods work only for one specific application. The difficulties arise because the geometry of the line networks shows various shapes. Also, there is often insignificant contrast between a pixel belonging to a curvilinear structure and the background texture. Thus, the information from an individual pixel fails to interpret the such topology. 
\begin{figure}[t]
	\centering
	\subfigure[Input] {\includegraphics[width=2cm]{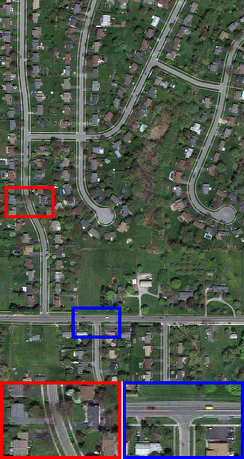}}
	\subfigure[Segmentation] {\includegraphics[width=2cm]{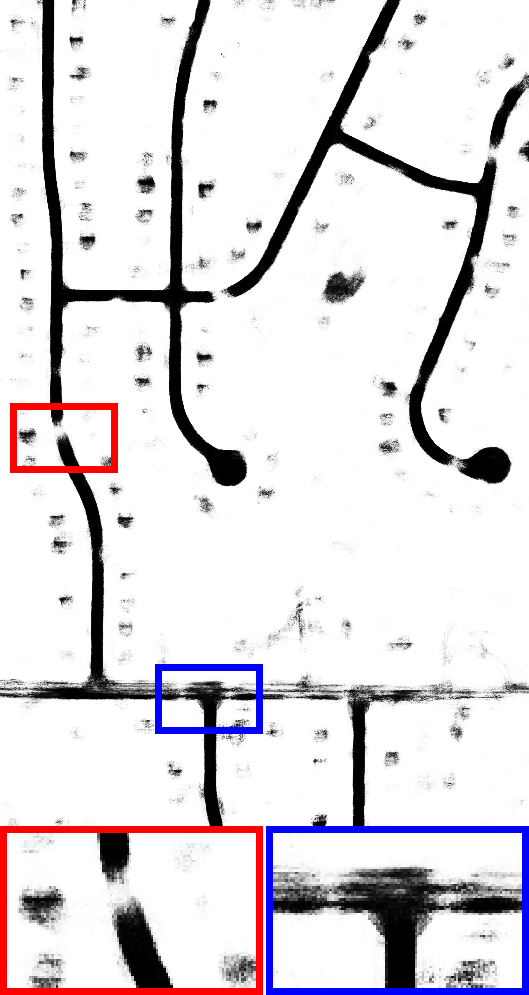}} 
	\subfigure[Centerline] {\includegraphics[width=2cm]{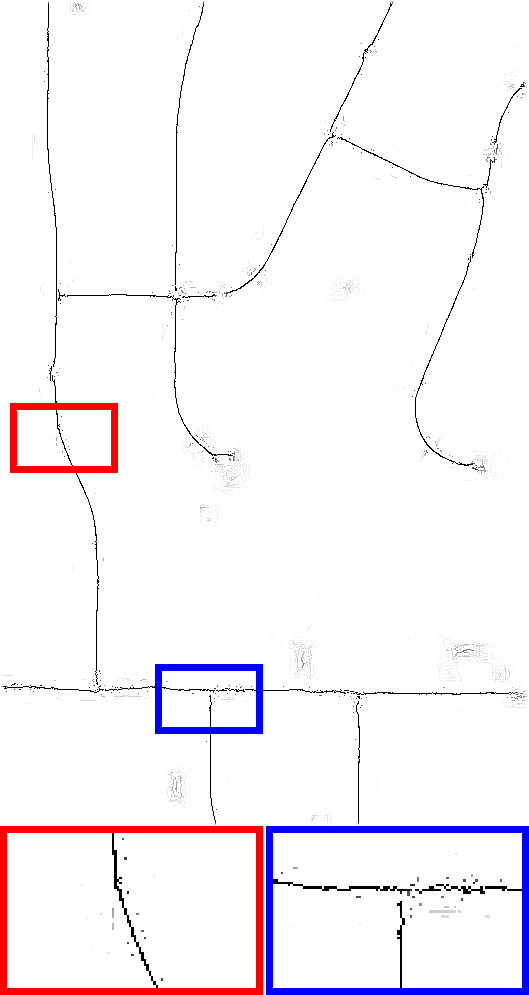}}
	\subfigure[Proposed] {\includegraphics[width=2cm]{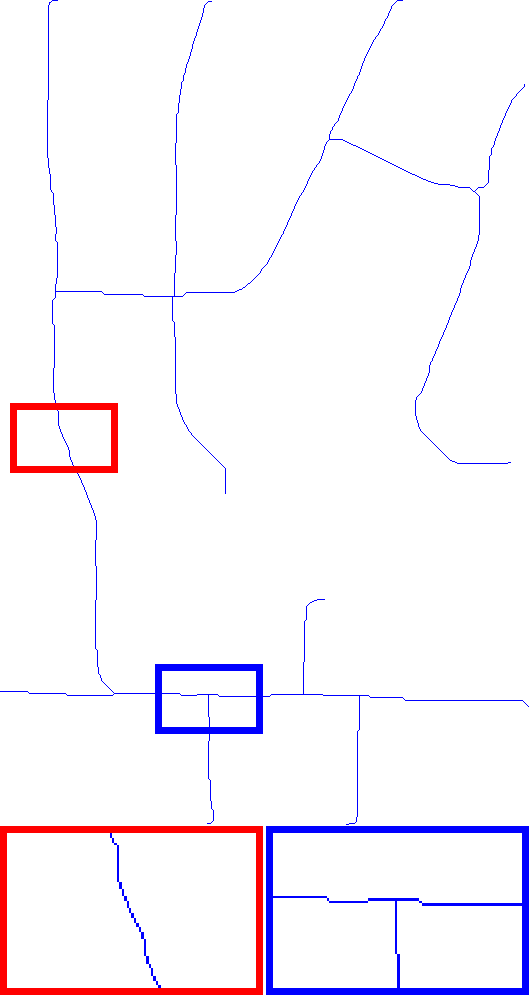}}					
	\caption{ 
		Compared with (b) the segmentation \cite{Becker2013} and (c) the centerline detection \cite{Sironi2014} methods, (d) the proposed algorithm represents well topological features of the curvilinear structure. In this example, road network is partially occluded by trees or cars. Setting a threshold value for the local measure often fails to detect the underlying curvilinear structure because this process neglects correlated information of the reconstructed curvilinear structure. Although the centerline is able to quantify scale (width) of curvilinear structure, it inaccurately classifies pixels around intersections. In this work we learn structured output ranking of the image patches that contain a part of curvilinear structures. The proposed graph-based representation uses the smallest pixels but also preserves the topological information.
	}\label{fig:concept}
\end{figure}
	To represent an arbitrary shape of the curvilinear structure, segmentation algorithms~\cite{Becker2013,Frangi1998,Law2008} evaluate linearity scores, which correspond to likelihood of pixels given the curvilinear structure. Then, a threshold value is set to discard pixels showing low scores. This process implicitly ignores correlated information of the pixels on the curvilinear structure, so that it yields discontinuities on the reconstruction results. Centerline detection algorithm~\cite{Sironi2014} is able to encode the width of the curvilinear structure; however, it is weak for localizing junctions of the curvilinear structure. On the other hand, graph-based models~\cite{Gonzalez2010,Peng2011,Wang2011,Zhao2011} reconstruct the curvilinear structure with geodesic paths in a sparse graph. For an efficient computation, these algorithms usually exploit a subset of pixels which correspond to local maxima of the linearity score. Contrary to the previous methods, our approach initially infers the curvilinear structure based on the {\it structured ranking scores}, which are related to the spatial patterns of the latent curvilinear structure. We then search for the longest geodesic path on the graph and iteratively add fine branches to rebuild the curvilinear structure with minimal pixels. Figure~\ref{fig:concept} compares the results obtained with curvilinear structure segmentation~\cite{Becker2013}, centerline detection~\cite{Sironi2014}, and the proposed method.

	In this work we aim to learn structured rankings of the input image patches which may contain a part of curvilinear structures. In particular, we assign a high ranking score for the image patch if it shows plausible spatial pattern to the latent curvilinear structures. Since the curvilinear structures are arbitrarily oriented within in an image patch, we measure the local orientation of the image patch and rotate it with respect to the baseline orientation (align horizontally). We then explore geodesic paths in the graph built upon the structured ranking score map. The proposed algorithm can provide the topological importance level of the curvilinear structure. Our experiments demonstrate that the proposed algorithm localizes the curvilinear structure with a high accuracy compared to the state-of-the-art methods.

\subsection{Related work}
\label{ssec:related_work}

	Geometrically speaking, curvilinear structures are rotatable and elongated shape. Image gradient information~\cite{Frangi1998,Freeman1991,Jacob2004,Law2008,Perona1995} is useful to estimate local orientation and to describe the shape of the curvilinear structure. The local image features are insufficient to discern curvilinear structures from undesirable high-frequency components due to the lack of geometric interpretation. To take the structural information into account, graphical models~\cite{Gonzalez2010,Turetken2013} formulate an energy optimization problem with spatial constraints in a local configuration. Stochastic models~\cite{Jeong2015,Lacoste2005} also specify a distribution of the line objects given image data with pairwise interaction terms. A sampling technique~\cite{Green1995} is employed to maximize the probability density. However, the geometric constraints are heuristically designed, and the number of constraints are increased to describe complex shaped line networks. Recently, machine learning algorithms have been involved to detect curvilinear structures latent in various types of images. Becker~\etal~\cite{Becker2013} applied a boosting algorithm to obtain an optimal set of convolution filter banks. A regression model is proposed to detect centerlines by learning the scale (width) of the tubular structures with the non-maximum suppression technique~\cite{Sironi2014}.

	Structured learning system has been employed in image segmentation models based on random fields~\cite{Bertelli2011,Kim2014,Lucchi2013,Szummer2008}. More specifically, Structured Support Vector Machine (SSVM)~\cite{Tsochantaridis2005} is used to predict model parameters for inference of the structured information between input image space and output label space. Exploring all possible combinations of the labels in the output space is computationally intractable. Thus, the random field models define the pairwise relationship in a neighborhood system to enforce the labeling consistency. While such prior models based on random fields are successful to describe convex shaped objects, it is inefficient to detect thin and elongated shaped curvilinear objects. 
			
	The main contributions of this work are summarized as follows: 
\renewcommand{\labelitemi}{$\bullet$}
\begin{itemize}
\item We propose an orientation-aware curvilinear feature descriptor for the curvilinear structure inference;
\item We learn a ranking function which can infer spatial patterns of the curvilinear structure within image patches; and
\item We reconstruct the latent curvilinear structure based on graph topology with the structured output rankings.
\end{itemize}
	\begin{figure}[t]
		\centering
		\subfigure{\includegraphics[width=8cm]{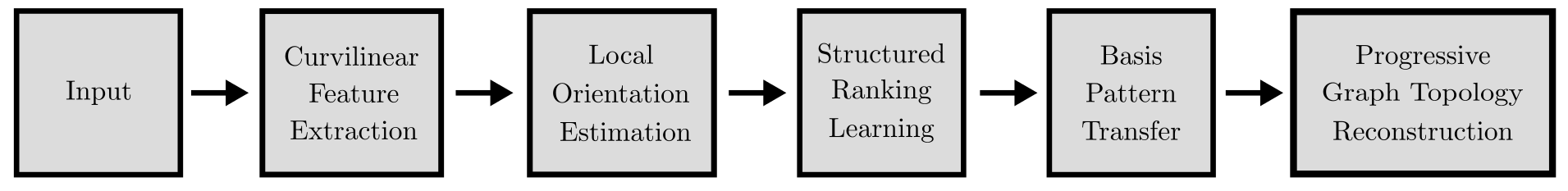}}    
		\caption{
		    Overview	 of the proposed algorithm
		}\label{fig:overview}
	\end{figure}
	The rest of the paper is organized as follows: Sec.~\ref{sec:method} provides the mathematical description and outline of our method. Sec.~\ref{sec:feature} proposes an orientation-aware curvilinear feature descriptor based on oriented image gradients. Sec.~\ref{sec:learning} explains how we infer the structured output rankings of the given input image patches. Sec.~\ref{sec:inference} develops a graph-based curvilinear structure reconstruction algorithm. Sec.~\ref{sec:results} shows experimental results on different types of datasets. Finally, Sec.~\ref{sec:conclusions} concludes this work.

	\begin{figure*}[ht]
    	\centering
    	\begin{tabular}{m{1.1cm}m{1.1cm}m{1.1cm}m{1.1cm}m{1.8cm}m{1.8cm}m{1.8cm}m{1.9cm}}  
    	& 
    	{\includegraphics[height=1.25cm]{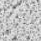}}&
    	{\includegraphics[height=1.25cm]{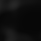}}& 
    	{\includegraphics[height=1.25cm]{figures/features/feat01.png}}&    	
    	{\includegraphics[width=1.8cm]{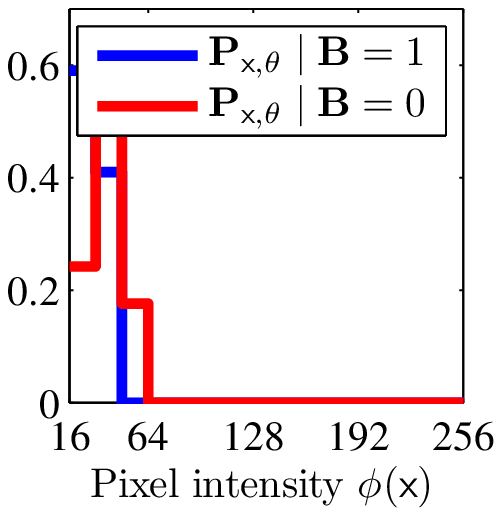}}& 
	{\includegraphics[width=1.8cm]{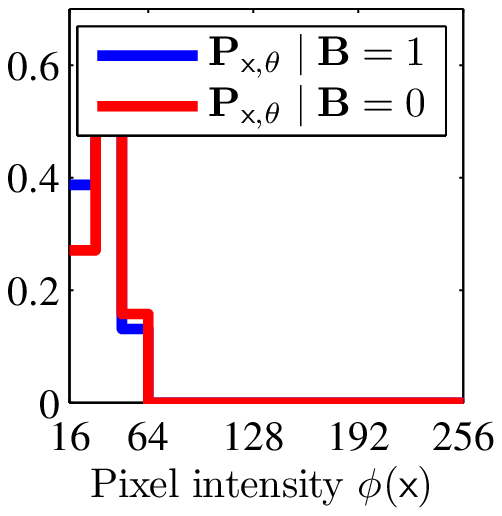}}& 
    	{\includegraphics[width=1.8cm]{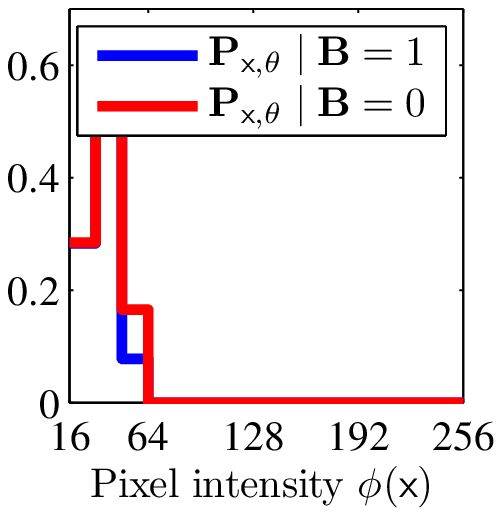}}& 
    	{\includegraphics[width=1.9cm]{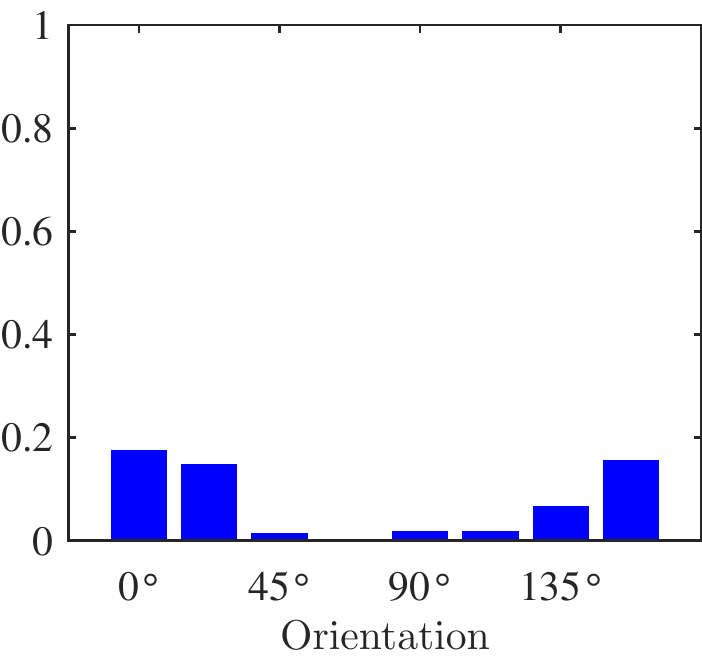}}\\[-0.25em]
    	
    	{\includegraphics[height=1.25cm]{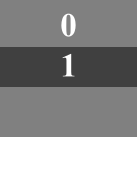}}&    	
    	{\includegraphics[height=1.25cm]{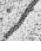}}&
    	{\includegraphics[height=1.25cm]{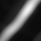}}&    	
    	{\includegraphics[height=1.25cm]{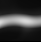}}&    	
    	{\includegraphics[width=1.8cm]{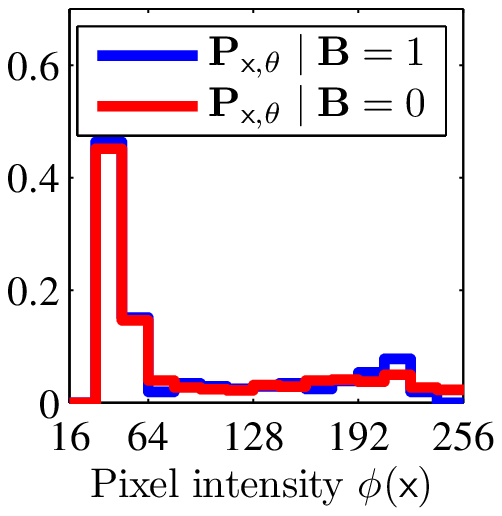}}& 
	{\includegraphics[width=1.8cm]{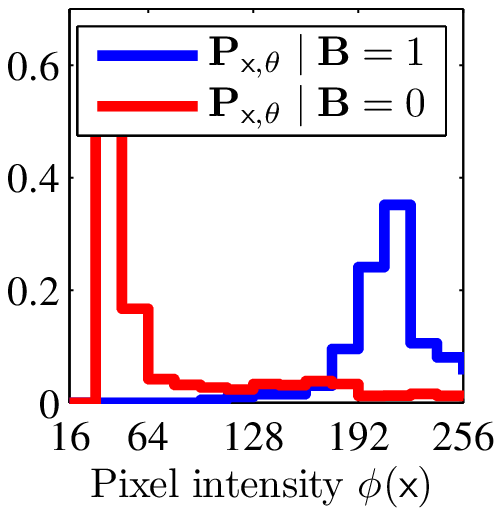}}& 
    	{\includegraphics[width=1.8cm]{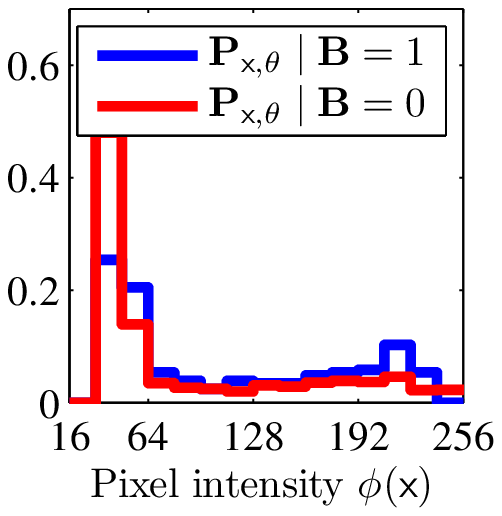}}& 
    	{\includegraphics[width=1.9cm]{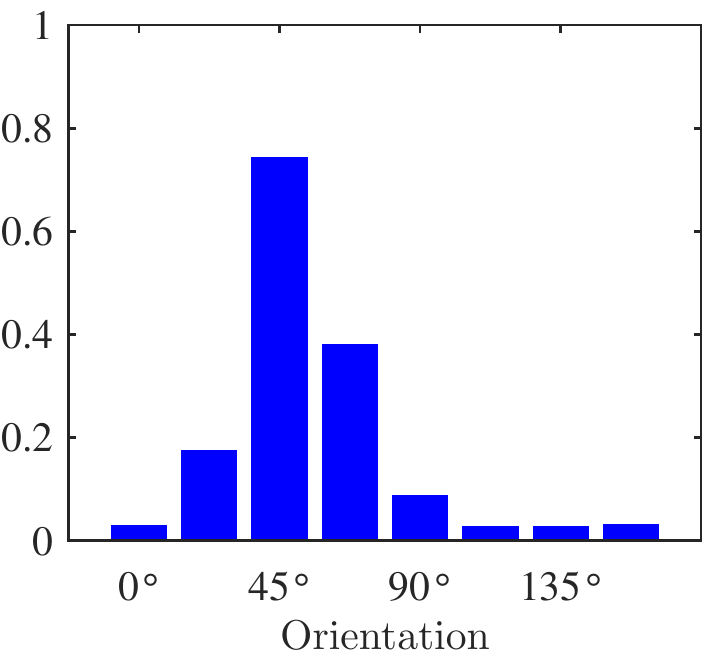}}\\[-0.25em] 
    	
    	\small\centering $\vect{B}$ & 
    	\small\centering $\bf{I}_{\pixel{x}}$ & 
    	\small\centering $\bf{P}_{\pixel{x}}$ & 
    	\small\centering ${\bf{P}}_{\pixel{x}, 45^\circ}$ & 
    	\small\centering $\theta=0^\circ$ & 
    	\small\centering $\theta=45^\circ$ & 
    	\small\centering $\theta=90^\circ$ & 
    \small\centering $\chi^2$
    	\end{tabular}  	    	
    	
		\caption{
		In this figure, we compare the statistics of pixel values in the rotated image patches with respect to the given binary mask $\vect{B}\in\{0,1\}^M$. Assume that $\vect{B}$ encodes a basis spatial pattern of the curvilinear structure toward the baseline orientation ($\bar\theta=0^\circ$), where darker gray color refers to $1$. To estimate local orientation of the input image patch, we rotate it and find the orientation which maximizes the statistical difference of pair distributions. If the input image patch does not contain curvilinear structure (upper row), there is no meaningful statistical difference for any orientation.
		}\label{fig:ori_estimation}	
	\end{figure*}
\section{Overview}
\label{sec:method}

	In this section, we define notations and provide an overview of the proposed algorithm (see Figure~\ref{fig:overview}). Assume that an image $I$ contains a curvilinear structure. We denote the latent curvilinear structure $\Omega:I\mapsto\{0,1\}$ for any pixel $\pixel{x}$ of the image $I$:
	\begin{equation}\label{eq:groundtruth}
		\Omega(\pixel{x}) = \left\{
		\begin{array}{ll}
			1	\quad & \mbox{\rm if $\pixel{x}$ is on the curvilinear structure},\\
			0	\quad & \mbox{otherwise}.\\
		\end{array}	\right.
	\end{equation}
This function is also related to a ground truth map, which is manually labeled, for the machine learning framework and for the performance evaluation. We compute a curvilinear feature map $\phi:I\mapsto\mathbb{R}$ that represents oriented gradient information (see Section~\ref{ssec:feature_extraction}). Since information embedded in a single pixel is limited to infer the latent spatial patterns, we exploit image patches to compute input feature vectors. Let ${\bf{P}}_{\pixel{x}}$ be a patch of the feature map values within $\sqrt{M}\times\sqrt{M}$ size of square window centered at $\pixel{x}$,~\ie,~${\bf{P}}_{\pixel{x}}=\{\phi(\pixel{x}')\mid\|\pixel{x}-\pixel{x}'\|_\infty\leq\tfrac{\sqrt{M}}{2}\}$. Using a rotation matrix $\vect{R}_\theta$ defined on Euclidean image space, we can rotate a patch with respect to the given orientation $\theta$ such as ${\bf{P}}_{\pixel{x},\theta}=\{\phi(\pixel{x}')\mid\|\vect{R}_\theta\tr(\pixel{x}-\pixel{x}')\|_\infty\leq\tfrac{\sqrt{M}}{2}\}$.
	
	We use graph theory for shape simplification of the complex curvilinear structure. For the pixels showing higher rankings, we build an undirected and weighted graph $G=(V,E)$. Then, we look for the longest geodesic path which corresponds to the coarset curvilinear structure in the image. We iteratively reconstruct the curvilinear by collecting paths that connect the remotest vertices in the graph. Consequently, the proposed algorithm can represent the different levels of detail in the latent curvilinear structure using the minimum number of pixels.

\section{Curvilinear Feature}
\label{sec:feature}
	This section is devoted to compute the curvilinear feature descriptor that is used for the inputs of the learning system. We perceive the latent curvilinear structure based on inconsistency of background textures and its geometric characteristics. In other words, a sequence of pixels corresponding to the curvilinear structure has different intensity values compared to its surroundings, and shows thin and elongated shape. Thus, we compute multi-direction and multi-scale image gradients to detect locally oriented image features.
	\subsection{Curvilinear feature extraction}
	\label{ssec:feature_extraction}
	We obtain oriented gradient maps using a set of steerable filters~\cite{Freeman1991,Jacob2004,Perona1995}. Before applying the convolution operations, we normalize the training and test images to remove the effects of various illumination factors: 
\begin{equation}\label{eq:normalization}
	\tilde{I} = \frac{1}{1+e^{-\tfrac{I-\mathbb{E}[I]}{\max(I)-\min(I)}}},
\end{equation}
where $\mathbb{E}[I]$ is the sample mean of the image.

	The steerable filters are created by the second order derivative of isotropic 2D Gaussian kernels. Let ${\bf f}_{\theta,\sigma}$ be a steerable filter that accentuates image gradient magnitude for direction $\theta$ at scale $\sigma$. To take into account varying orientations and widths of the curvilinear structure, a feature map $\phi$ is combined by multiple filtering responses:
	\begin{equation}
		\phi = \frac{1}{|\Theta|}\sum_\theta\max_\sigma\{{\bf f}_{\theta,\sigma}*\tilde{I}\},
	\end{equation}
where $|\Theta|$ denote the total number of orientations. We first find the maximum filtering responses of the scale spaces, and then average for all directional responses. In this work we consider 8 orientations and 3 scales. 


	\subsection{Local orientation estimation}
	\label{ssec:estimate_orientation}
	In the previous subsection, we evaluate the presence of curvilinear structure from the amount of gradient magnitudes in image patches. Complex shaped curvilinear structure also consists of various local orientations. 

	Assume that $\vect{B}\in\{0,1\}^M$ be a basis spatial pattern of the curvilinear structure for the baseline orientation ($\bar\theta=0^\circ$). 	We manually design $\vect{B}$ as a simple line shape template which highlights the middle of image patch (see Figure~\ref{fig:ori_estimation}). If an image patch ${\bf{P}}_{\pixel{x}}$ contains a part of curvilinear structure, pixels on the curvilinear structure are sequentially accentuated towards a particular direction $\theta$. Recall that the geometric properties of the curvilinear structure, it is rotatable and symmetric. We can rotate image patch to be aligned with the basis spatial pattern.
		
	More specfically, we estimate the local orientation shown in image patches using statistical difference of the areas determined by $\vect{B}$. For a rotated image patch ${\bf{P}}_{\pixel{x},\theta}$, we compute normalized histograms of the pixels which are labeled as curvilinear structure $p_{\pixel{x},\theta}=\hist{{\bf P}_{\pixel{x},\theta}\mid\vect{B}}$ and its counterpart $q_{\pixel{x},\theta}=\hist{{\bf P}_{\pixel{x},\theta}\mid\neg\vect{B}}$, where $\hist{\cdot}$ denotes histgram of the given distribution with $32$ bins. Similar to~\cite{Arbelaez2011}, we employ ${\mathop{\chi}}^2$ test to compute statistical difference of two distributions: 
	\begin{gather}
		\hat{\theta} = \argmax_{\theta}{\mathop{\chi}}^2\left(p_{\pixel{x},\theta},q_{\pixel{x},\theta}\right),\label{eq:chi}\\
		{\mathop{\chi}}^2(p,q)=\frac{1}{2}\sum_i\frac{\left(p(i)-g(i)\right)^2}{p(i)+g(i)}.
	\end{gather}	
Figure~\ref{fig:ori_estimation} compares statistics of the image patches with respect to different orientations. If an image patch contains curvilinear structure, $\chi^2$ test score shows uni-modal distribution which shows a peak at the dominant orientation.

\section{Learning}
\label{sec:learning}
	We regulate the orientation and the shape of the input patches with respect to the basis spatial pattern $\vect{B}$. In this section, we aim to learn a function that predicts structured output rankings of the input image patches. Thus, the output rankings indirectly infer the spatial patterns of the curvilinear structure for image patch. 

	Let $\opvec{\cdot}$ be an operator to convert an image patch into a column vector. For a pixel $\pixel{x}_i$, the input feature vector is defined as $\vect{z}_i=\opvec{{\bf P}_{\pixel{x}_i,\theta_i}\mid\vect{B}}\in\mathbb{R}^N$ and $y_i\in\mathbb{R}^+$ denote the corresponding rank value. For a feature vector, we exploit subsampled pixels on the linear template to avoid data imbalance. Thus, the dimension of feature vector $N$ is small than the total number of pixels within the patch, \ie, $N\leq M$. 

	For the setup of machine learning, a training dataset $\mathcal{D}=\{(\vect{z}_i,y_i)\}_{i=1}^K$ consists of the $K$ input-and-output pairs. Let $\{\vect{z}_1,\dots,\vect{z}_K\}\in\mathcal{Z}$ be an unordered list of the input feature vectors and $\{y_1,\dots,y_K\}\in\mathcal{Y}$ be the corresponding classes. The input feature vector $\vect{z}$ is built from the curvilinear feature map $\phi$. Our goal is to learn a ranking function $h(\vect{z})$ which assigns a global ordering (ranking) of feature vectors: $h(\vect{z}_i) > h(\vect{z}_j) \Leftrightarrow y_i > y_j$. The inner product of the model parameter and a feature vector $\sv\tr\vect{z}$ is used to predict ranking score of the given image patch.
	
	In our work {\bf Structured SVM} framework~\cite{Joachims2006,Mittal2012} is employed to exploit the structure and dependencies within the output space $\mathcal{Y}$. To encode the structure of the output space, a loss value $\Delta$ of each input feature vector is defined: $h(\vect{z}_i)>h(\vect{z}_j)\Leftrightarrow\Delta_i<\Delta_j$. The loss value evaluates the quality of the learning system. Intuitively, the input image patches containing curvilinear structure are comparable to the basis spatial pattern. Hence, as an image patch is similar to the basis spatial pattern, we should put it on a higher ranking. Specifically, we compute the loss value as the overlapping ratio between image patch from the groundtruth map $\Omega$ and the basis $\bf B$:
	\begin{equation}\label{eq:sim}
		\Delta_i=1-\frac{|{\bf\Omega}_{\pixel{x}_i,\theta_i}\cap\vect{B}|}{|{\bf\Omega}_{\pixel{x}_i,\theta_i}\cup\vect{B}|},
	\end{equation}	
where ${\bf\Omega}_{\pixel{x}_i,\theta_i}$ denotes a rotated image patch referring to groundtruth map $\Omega$. The objective function of Structured SVM with a single slack variable $\slack$ is given by:
	\begin{align}\label{eq:svm_struct}
	\begin{array}{c l}
		\displaystyle\min_{\sv,\slack\geq 0}  & \frac{1}{2}\sv\tr\sv + C\slack  \\
		\textrm{s.t.}             & \frac{1}{|\mathcal{N}|}{\sum\limits_{(i,j)\in\mathcal{N}}}c_{ij}\sv\tr(\vect{z}_i-\vect{z}_j) \geq  \frac{1}{|\mathcal{N}|}\sum\limits_{(i,j)\in\mathcal{N}}c_{ij}-\frac{\slack}{\Delta_j-\Delta_i},\\
		& \forall(i,j)\in\mathcal{N}, ~\forall c_{ij}\in\{0,1\}\\
	\end{array}	
	\end{align}
where $c_{ij}$ is the indicator variable to reduce the number constraint in linear complexity $O(K)$:
	\begin{equation}
	c_{ij} = \left\{
	\begin{array}{ll}
		1      \quad & {\rm if~} \Delta_i < \Delta_j {\rm ~and~} \sv\tr\vect{z}_i-\sv\tr\vect{z}_j<1,\\
		0  		\quad & \mbox{otherwise}.\\
	\end{array}	\right. 
	\end{equation}

	{\it Cutting plain algorithm}~\cite{Tsochantaridis2005} is employed to solve (\ref{eq:svm_struct}). For the details about the Structured SVM optimization, we refer the reader to~\cite{Joachims2006,Mittal2012,Tsochantaridis2005}. In the following section, we plot the initial segmentation map from the structured rankings scores. We also exploit this ranking scores for the shape simplification based on graph traversal algorithm. 

\section{Curvilinear Structure Reconstruction}
\label{sec:inference}
\begin{figure*}[t]
		\centering
		\subfigure[]{\includegraphics[width=2.76cm]{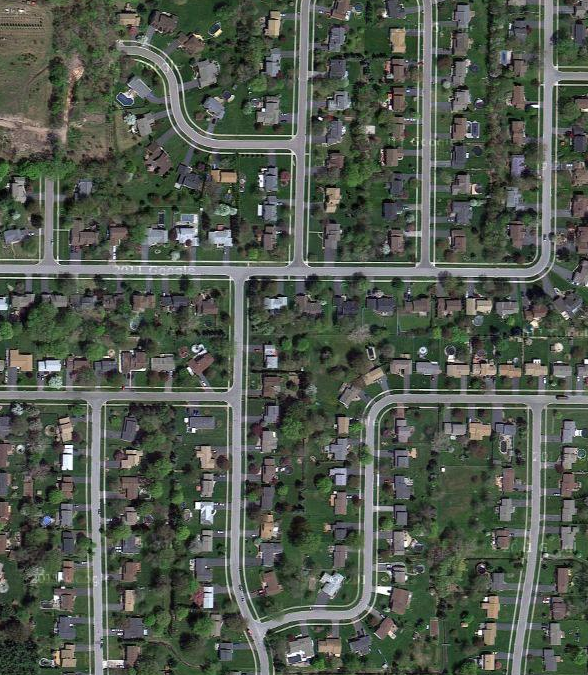}}    
		\subfigure[]{\includegraphics[width=2.76cm]{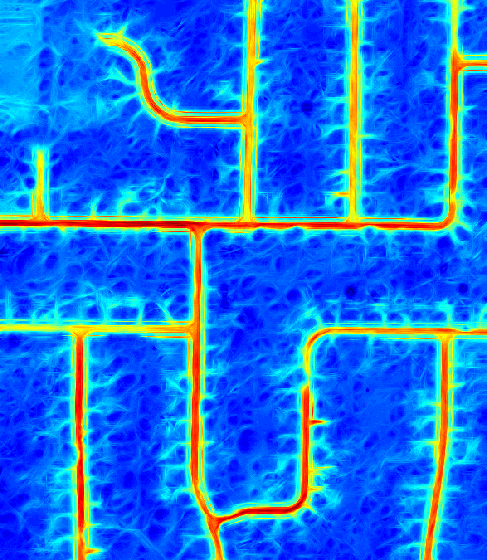}}
		\subfigure[]{\includegraphics[width=2.76cm]{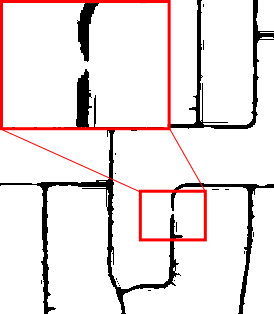}}
		\subfigure[]{\includegraphics[width=2.76cm]{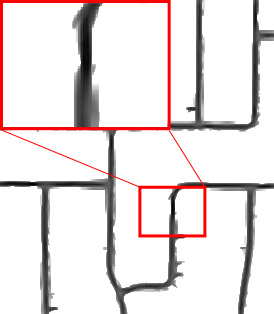}}
		\subfigure[]{\includegraphics[width=2.76cm]{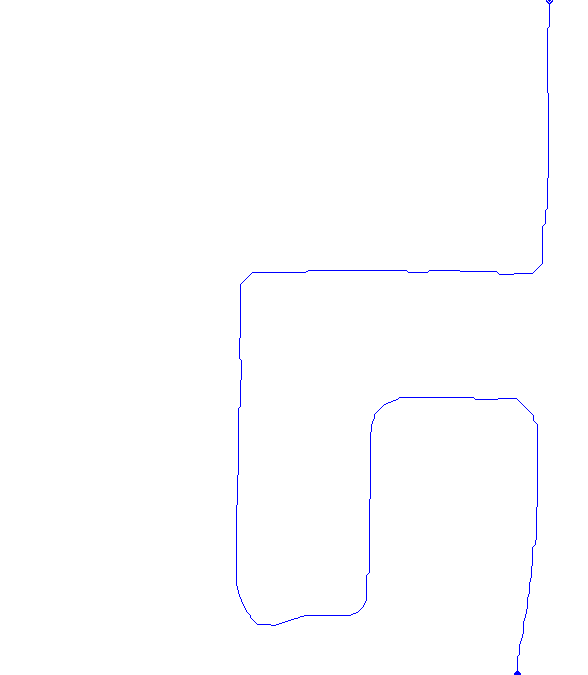}}	
		\subfigure[]{\includegraphics[width=2.76cm]{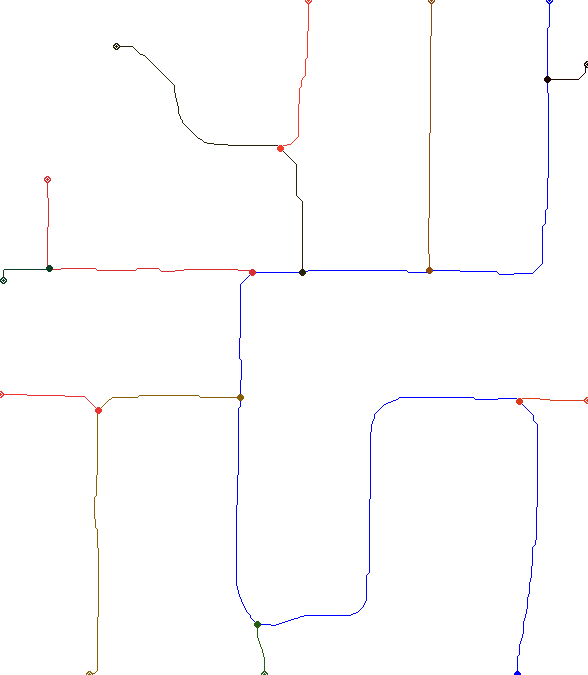}}	
		\caption{
			For the given image (a), we compute (b) the structured output ranking with Structured SVM. We retain binary map (c) from the highest to $\rho|I|$-th rankings. However, it could be broken. We exploit the basis spatial pattern $\bf B$ to reconstruct (d) the structured score map. Graph-based representation provides a tool to analyze the topological features of the curvilinear structures. (e) The coarsest curvilinear structure evolves to define (f) branches.
		}\label{fig:representations}
	\end{figure*}

	To represent the latent curvilinear structure, various local measures have been proposed to detect irregular shaped curvilinear structure. The score of the local measure is regarded as a likelihood probability whether a pixel is on the latent curvilinear structure. Most of the previous works represent the curvilinear structure as a binary map based on the linearity scores of the pixels. This approach easily misinterpret the topological features. In this section, we propose a novel structured score map based on the output ranking scores and the basis spatial pattern. We also develop a graph-based model which is able to organize the topological features of the structure in different levels of detail. Figure~\ref{fig:representations} shows step-by-step processing results to reconstruct the latent curvilinear structure.

	\subsection{Structured score map}
	\label{ssec:segmentation}
	We have obtained a score function $h(\vect{z};\sv)$ that evaluates the compatibility between the input features and the underlying curvilinear structure based on the structured output rankings. From the training dataset, we pre-compute the proportion $\rho$ of pixels being part of the curvilinear structure. Note that the value $\rho$ maximizes $F_1$ test scores at the groundtruth maps in the training dataset. During the test phase, we retain the output ranking according to the output ranking scores from the highest to the $\rho|I|$-th rank. As shown in Figure~\ref{fig:representations} (c), a binary map can be disconnected.
		
	To avoid unwilling breakpoints, we composite the structured score map $\Pi:I\mapsto\mathbb{R}$ using the binary map, induced by ranking scores $\sv\tr\vect{z}$, and the basis spatial pattern $\vect{B}$. Recall that we initially estimate the ranking score using the shape similarity between oriented image patch of ground truth and the basis spatial pattern in (\ref{eq:sim}). Thus, it can be regarded as the inverse mapping from output ranking score to the latent curvilinear structure. Let $\vect{b}=\opvec{{\bf B}}$ be the column vector version of the binary mask $\vect{B}$ and ${\bf Q}_{\pixel{x}_i,\theta_i}$ be the patch of structured score map $\Pi$ centered at $\pixel{x}_i$ with $\theta_i$. We minimize the following cost function with respect to $\boldsymbol{\pi}_i=\opvec{{\bf Q}_{\pixel{x}_i,\theta_i}}$:
	\begin{equation}
		J=\min\sum_{\pixel{x}_i\in I'}\|\vect{b}-(\sv\tr \vect{z}_i)\boldsymbol{\pi}_i\|_2^2.
	\end{equation}
This is a least square problem, so that the solution is found at points which satisfies $\frac{\partial J}{\partial\boldsymbol{\pi}}=0$. To reconstruct the structured segmentation map, we synthesize obtained patches ${\bf Q}_{\pixel{x}_i,\theta_i}=\opvecinv{\hat{\boldsymbol{\pi}}_i}$ at the subsampled grid points on image $\pixel{x}_i\in I'$, where $\hat{\boldsymbol{\pi}}_i=(\sv\tr \vect{z}_i)^{-1}\vect{b}$. We refer the readers to~\cite{Kwatra2005} for the implementation details of the related texture synthesis technique.
\begin{algorithm}[t]
	\small
	\caption{Progressive curvilinear path reconstruction}
	\label{algo:algorithm}
	\begin{algorithmic}[1]
	\State{\bf Inputs}: 
	\NoNumber{~~\quad$G'=(V',E')\sim$ a subgraph of $G$; and}	
	\NoNumber{~~\quad$\hat\ell \sim$ a minimum length of the curvilinear structure}
	\State{\bf Output}:
	\NoNumber{~~\quad$P \sim$ a set of vertices corresponding to the simplified curvilinear structure}

	\State $P \leftarrow \emptyset$
	\State {\tt longest\_path\_length} $\leftarrow |V'|$ 
	\While{}	
		\State Compute the longest geodesic path $T$ in $G'$ using~\cite{Corneil2001}
		\State {\tt longest\_path\_length} $\leftarrow |T|$
		\If {{\tt longest\_path\_length}$ <\hat\ell$}
			\State break
		\EndIf
		\State $P \leftarrow P\cup T$ 
		\State $w(\pixel{u},\pixel{v})\leftarrow 0,~ \forall\{\pixel{u},\pixel{v}\}\in T$, Update all edge weights on the path $T$ as $0$ for the given subgraph $G'$
	\EndWhile
	\end{algorithmic}
\end{algorithm}

	\subsection{Progressive curvilinear path reconstruction}
	\label{ssec:graph_representation}
	We consider the graph $G=(V,E)$ where $V$ is the set of pixels. Two pixels $\pixel{u}$ and $\pixel{v}$ are connected by the edge if and only if $\|\pixel{u}-\pixel{v}\|\leq \sqrt{2}$. Moreover, we assign a weight for each edge $\{\pixel{u},\pixel{v}\}\in E$ as $w(\pixel{u},\pixel{v})=\|\pixel{u}-\pixel{v}\|\frac{\Pi({\pixel{u})+\Pi(\pixel{v}})}{2}$. A path $P$ in the graph $G$ is a sequence of distinct vertices such that consecutive vertices are adjacent. The length of $P$ is the sum of the weights of its edges, and the distance $\dist{\pixel{u},\pixel{v}}$ between two vertices $\pixel{u}$ and $\pixel{v}$ is the minimum length of a path from $\pixel{u}$ to $\pixel{v}$. The eccentricity $\ecc{\pixel{v}}$ denotes the maximum distance from the vertex $\pixel{v}$ to a vertex $\pixel{u}\in V$, \ie,~$\ecc{\pixel{v}}=\max_{\pixel{u}\in V}\dist{\pixel{u},\pixel{v}}$. The diameter $\diam{G}$ of $G$ equals $\max_{\pixel{v}\in V} \ecc{\pixel{v}}$, \ie,~it is the maximum distance between two vertices in $G$. 

	To simplify the latent curvilinear structure, we look for long geodesic paths in the subgraph $G'$ of $G$ induced by the pixels with structured score map $\Pi$. More precisely, our algorithm computes a diameter of $G'$, \ie,~a shortest path $T$ with length $\diam{G'}$. This path $T$ is added in the simplified curvilinear structure, then the path $T$ is contracted into a single (virtual) vertex. For the implementation, the weight of all edges of $T$  become $0$. We repeat this process till the diameter of the subgraph is larger than pre-defined path length $\hat\ell$. The entire procedure is summarized in Algorithm~\ref{algo:algorithm}.

\begin{figure}[t]
\centering
	\subfigure[]{\includegraphics[width=2.4cm]{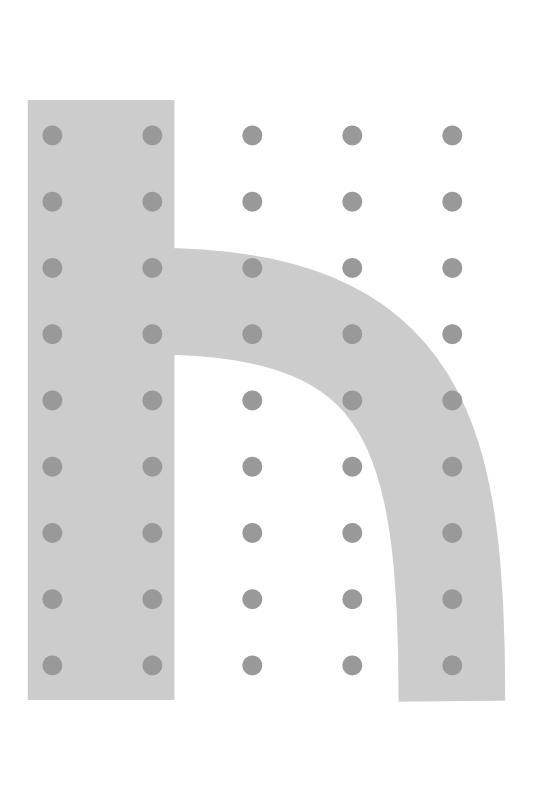}}
	\subfigure[]{\includegraphics[width=2.4cm]{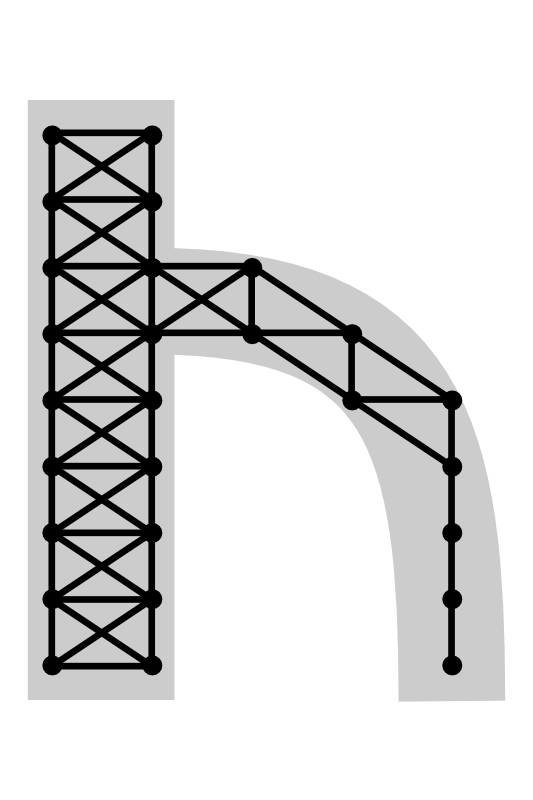}}
	\subfigure[]{\includegraphics[width=2.4cm]{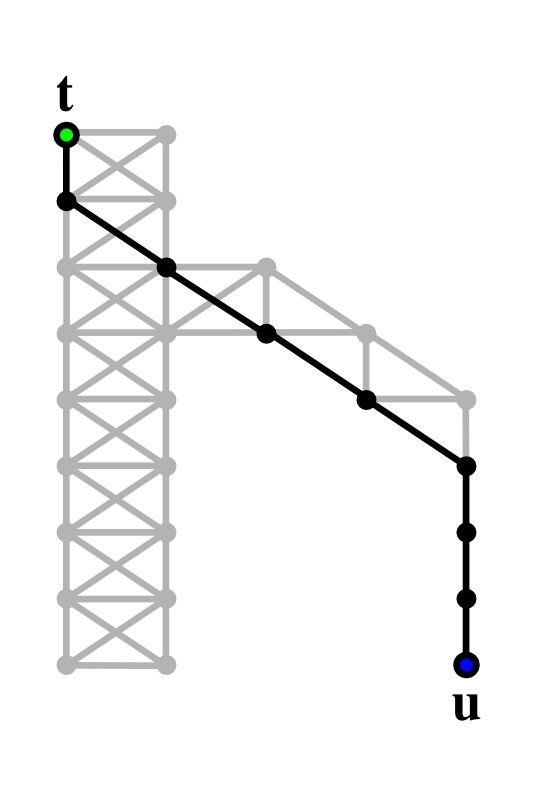}} \\
	\subfigure[]{\includegraphics[width=2.4cm]{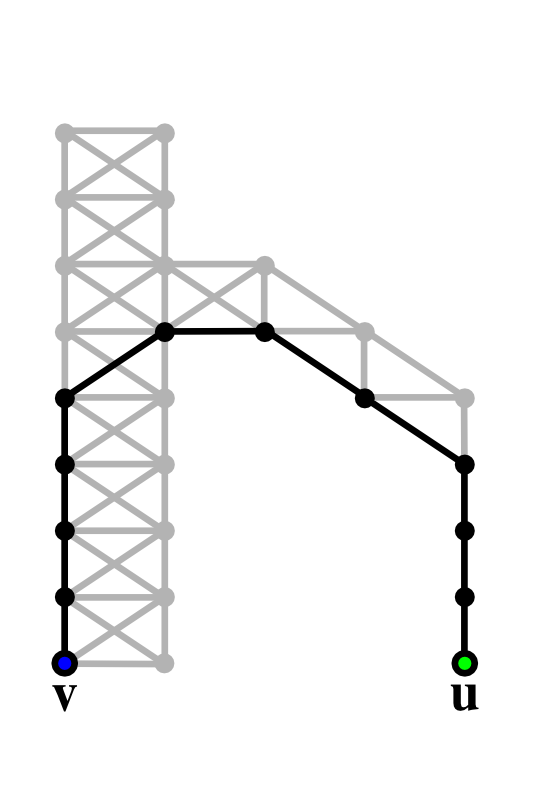}}
	\subfigure[]{\includegraphics[width=2.4cm]{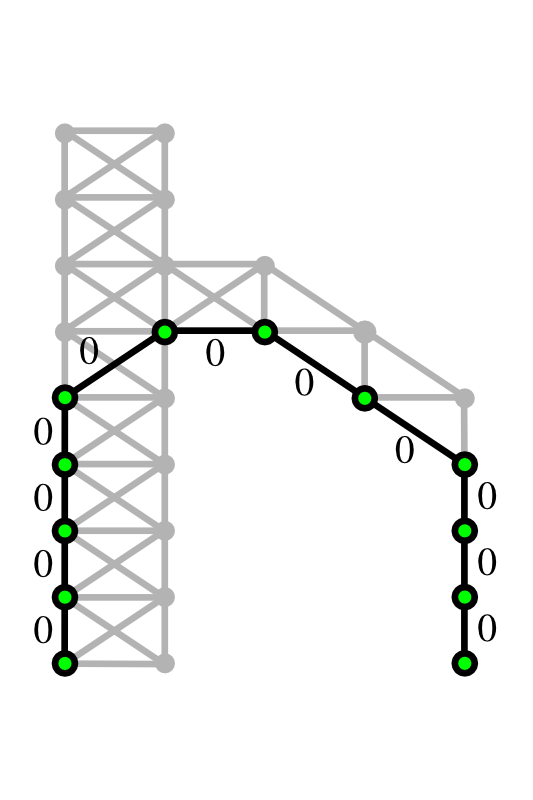}} 
	\subfigure[]{\includegraphics[width=2.4cm]{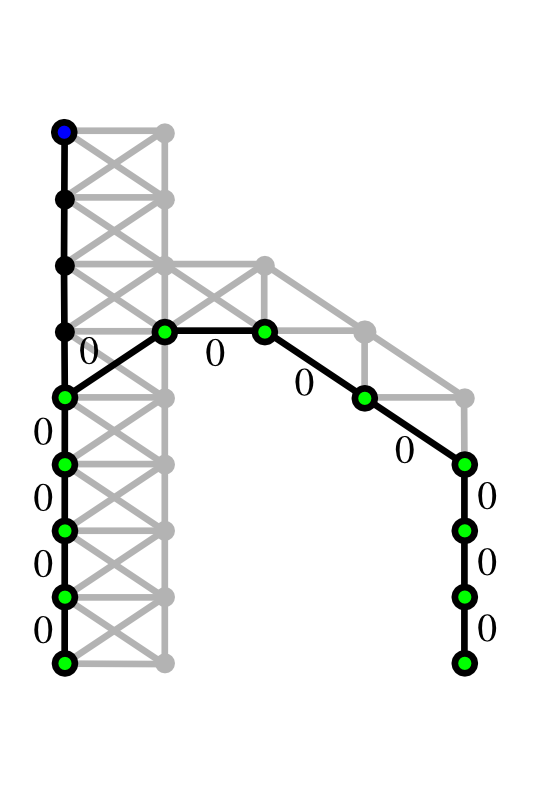}}
    \caption { 
    Toy example of the proposed curvilinear structure inference algorithm: (a) input image contains a curvilinear structure which is denoted by gray color; (b) subgraph $G'$ is induced from the structured segmentation map; (c) and (d) show the intermediate processes of the 2-sweep algorithm starting from vertex $\vect{t}$ to find a diameter of the subgraph; (e) we assign $0$ weight for all edges on the path; and (f) we repeat the process and add branches if the path length is larger than pre-defined length $\hat\ell$.
	}\label{fig:process}
\end{figure}

	The time-consuming part of the proposed algorithm is the computation of a diameter of $G'$. Rather than computing all pair distances (which requires a linear number of application of Dijkstra's algorithm), we use an efficient heuristic algorithm called {\it 2-sweep algorithm}~\cite{Corneil2001}. The 2-sweep algorithm randomly picks a vertex $\pixel{t}$ in $G'$, then performs Dijkstra's algorithm from $\pixel{t}$ to find a vertex $\pixel{u}$ at the maximum distance from $\pixel{t}$, \ie,~$\dist{\pixel{u},\pixel{t}}=\ecc{\pixel{t}}$. Then, it computes (using Dijkstra's algorithm) a path from $\pixel{u}$ to a vertex $\pixel{v}$ at the maximum distance from $\pixel{u}$. The length of the second path (from $\pixel{u}$ to $\pixel{v}$) is a good estimation of $\diam{G'}$. Note that this algorithm is able to compute the exact diameter if the graph has tree structure~\cite{Borassi2015}. Topologically speaking, most of latent curvilinear structures are very close to trees~\cite{Gonzalez2010,Turetken2011}. Therefore, the 2-sweep algorithm is well adapted to reconstruct the tree-like curvilinear structures. For a better understanding, we schematically explain the intermediate steps of the proposed curvilinear structure simplification algorithm in Figure~\ref{fig:process}. 

\section{Experimental results}
\label{sec:results}
	In this section, we first discuss the parameters of the proposed algorithm and datasets. We then compare the quantitative and qualitative results of the proposed algorithm and those of competing models proposed by \cite{Frangi1998}, \cite{Law2008}, \cite{Becker2013}, and \cite{Sironi2014}.

	\subsection{Parameters and Datasets}
	The proposed algorithm requires few parameters to compute the curvilinear feature descriptor $\phi$. We use 8-different orientations in this work: $\Theta=\{0^\circ,22.5^\circ,45^\circ,\\57.5^\circ,90^\circ,112.5^\circ,135^\circ,157.5^\circ\}$. The size of steerable filters is fixed to $21\times 21$ pixels. The scale factors $\sigma^2$ and the minimum path length $\hat{\ell}$ are adaptively selected for each dataset to obtain the best performances. The binary spatial pattern $\vect{B}$ is based on the physical attribute of the dataset in that  thickness $\tau$ of curvilinear structure shows various depending on the dataset. For Structured SVM training, we sample 2000 image patches on each dataset. All patch sizes are fixed to $M=33^2$. $C$ which controls the relative importance of slack variables is set to $0.1$ for all datasets. We choose the set of parameters for the SSVM training via 3-fold cross validation~\cite{Kohavi1995} which maximizes the average $F_1$ score~\cite{Martin2004} of the training set. 

	We test our curvilinear structure model on the following public datasets:
	
\renewcommand{\labelitemi}{$\bullet$}
	\begin{itemize}
	\item {\bf DRIVE}~\cite{Staal2004}: The dataset consists of 40 retina scan images with manual segmentation by ophthalmologists to evaluate the blood vessel segmentation algorithms. We use 20 images for the training and 20 images for the test, respectively. The path length $\hat{\ell}$ is set to $40$. The scale factors and thickness are set to $\sigma^2=\{2,4,8\}$ and $\tau=5$, respectively.
	\item {\bf RecA}~\cite{Jeong2015}: We collect electron microscopic images of RecA proteins on DNA which contain filament structure. We use 4 training images and 4 test images. We use $\hat{\ell}=30$, $\sigma^2=\{4,8,12\}$, and $\tau=5$.
	\item {\bf Aerial}~\cite{Sironi2014}: The dataset contains 14 remote sensing images of road networks. We select 7 images for the training and 7 images for the test, respectively. We use $\hat{\ell}=80$, $\sigma^2=\{4,8,12\}$, and $\tau=9$.
	\item {\bf Cracks}~\cite{Chambon2010}: Images of the dataset correspond to road cracks on the asphalt surfaces. We use 6 images to train and test the algorithms on different 6 images. For this dataset, we set to $\hat{\ell}=30$, $\sigma^2=\{2,4,8\}$, and $\tau=3$, respectively.
	\end{itemize}

	\subsection{Evaluations}
\begin{figure*}[t]
	\centering
		
	\subfigure[GT]
	{\includegraphics[width=2.3cm]{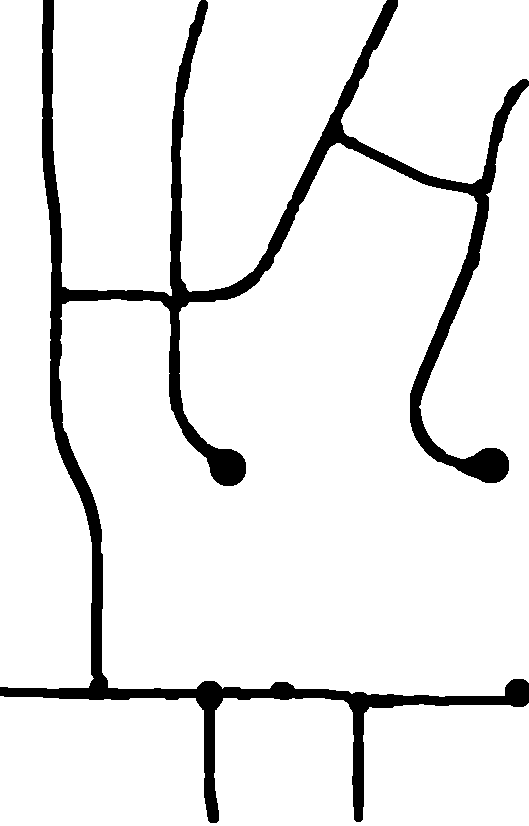}}
	\subfigure[\# 1]
	{\includegraphics[width=2.3cm]{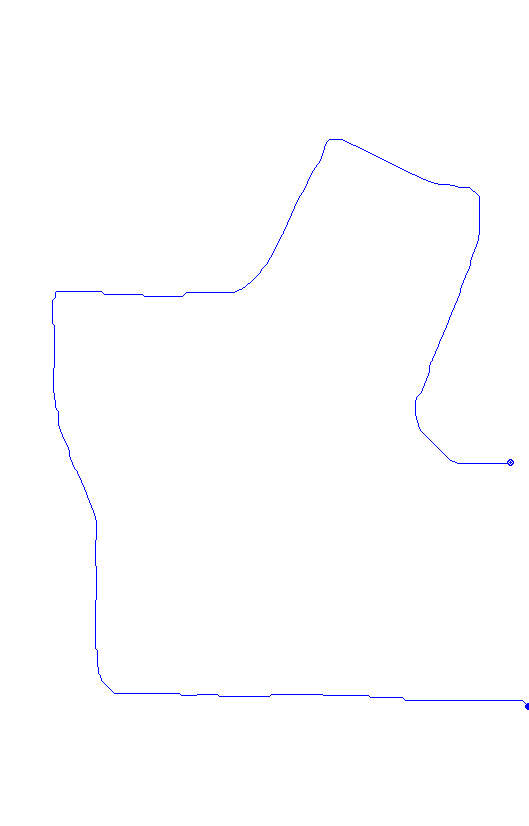}}
	\subfigure[\# 2]
	{\includegraphics[width=2.3cm]{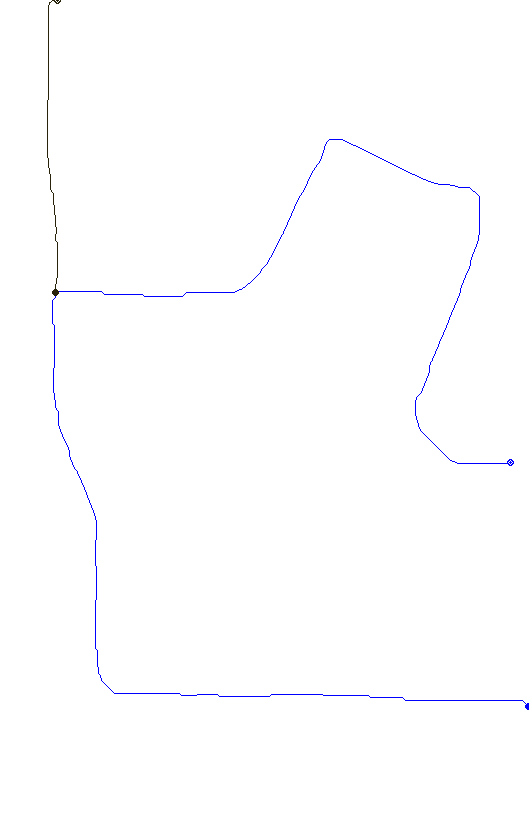}} 
	\subfigure[\# 3]
	{\includegraphics[width=2.3cm]{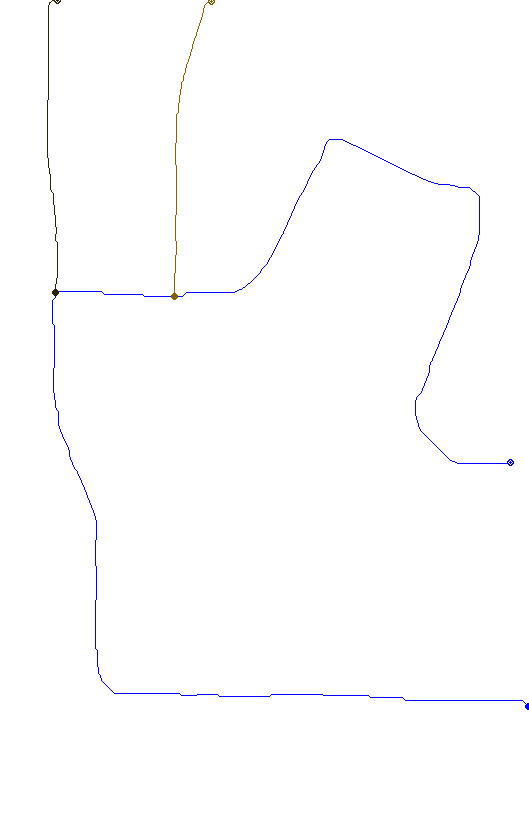}} 
	\subfigure[\# 4]
	{\includegraphics[width=2.3cm]{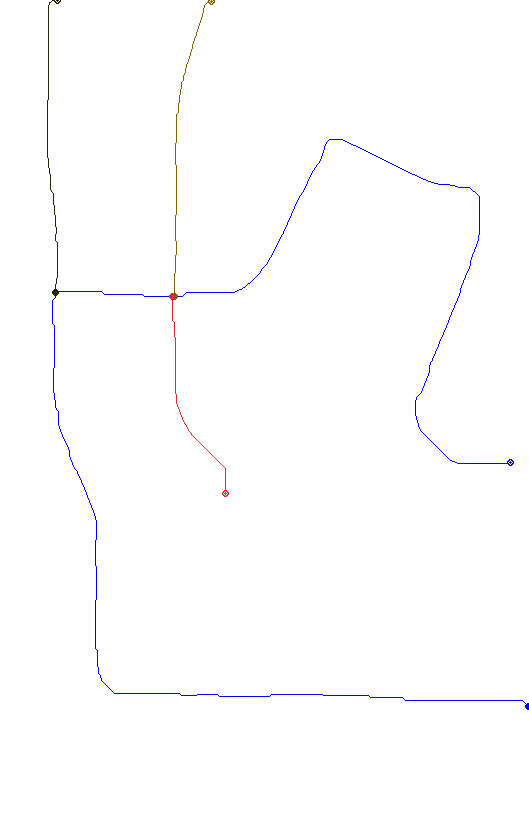}}
	\subfigure[\# 5]
	{\includegraphics[width=2.3cm]{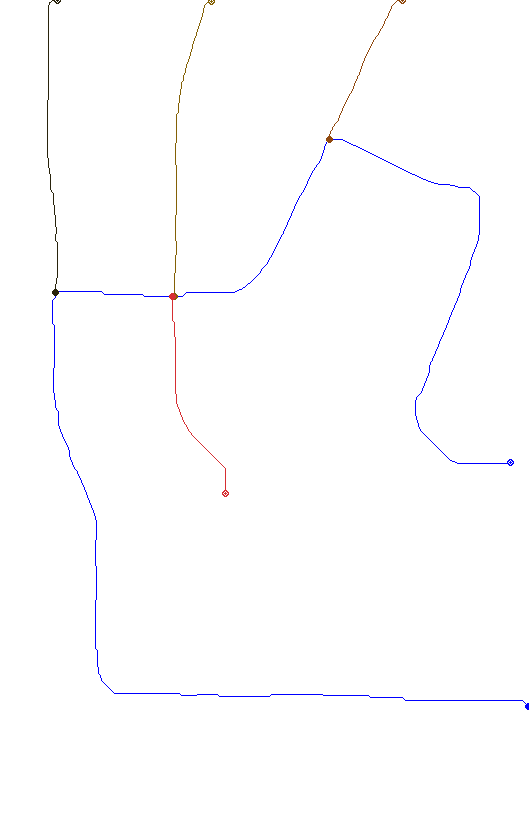}}
	\subfigure[\# 9]
	{\includegraphics[width=2.3cm]{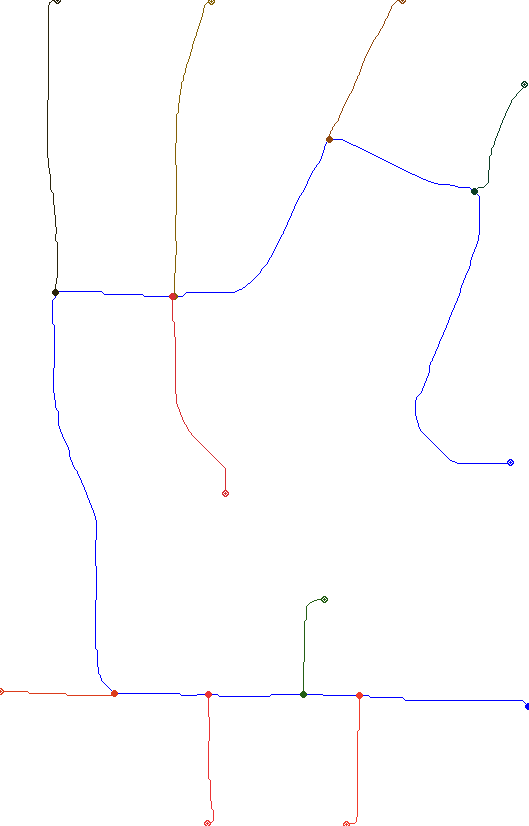}} 
	
	\subfigure[GT]
	{\includegraphics[width=2.3cm]{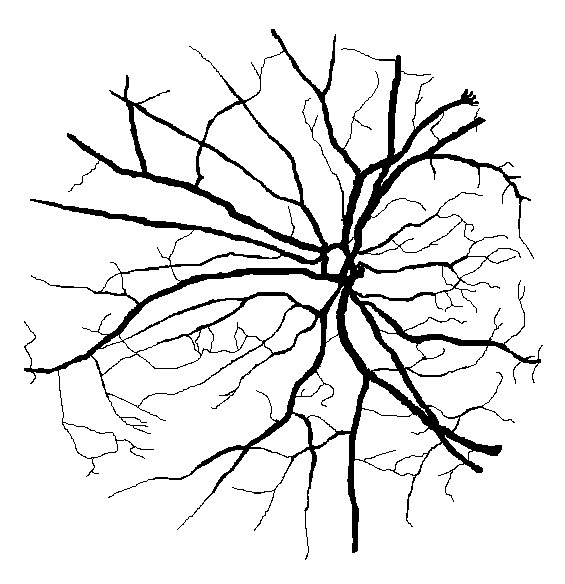}}
	\subfigure[\# 1]
	{\includegraphics[width=2.3cm]{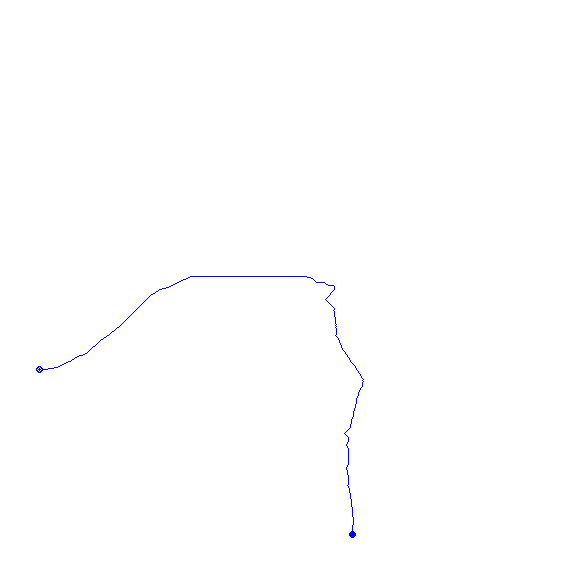}}
	\subfigure[\# 3]
	{\includegraphics[width=2.3cm]{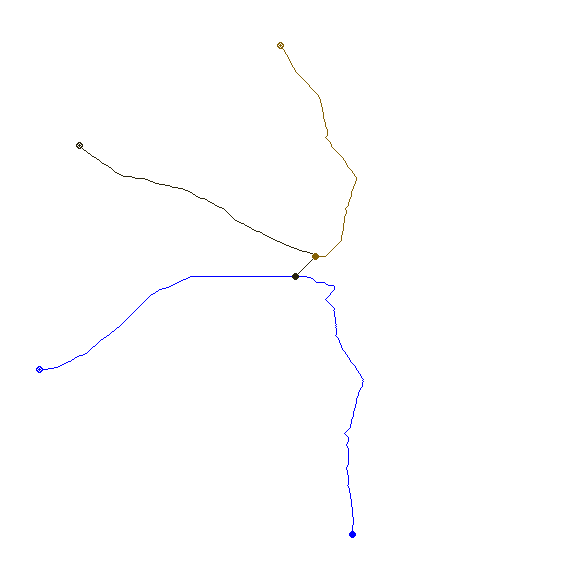}} 
	\subfigure[\# 5]
	{\includegraphics[width=2.3cm]{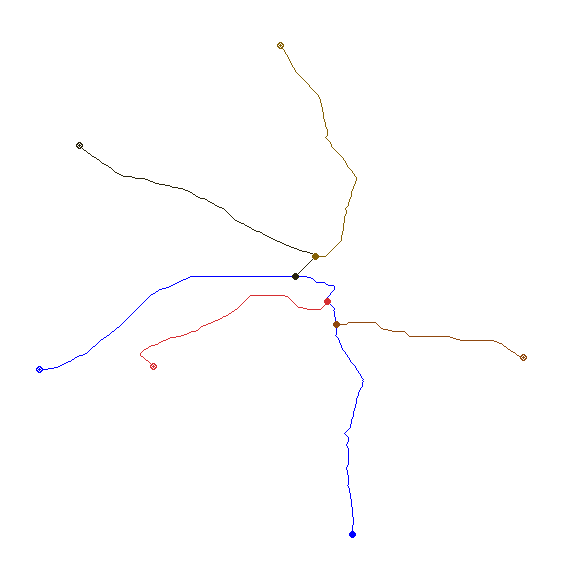}} 
	\subfigure[\# 7]
	{\includegraphics[width=2.3cm]{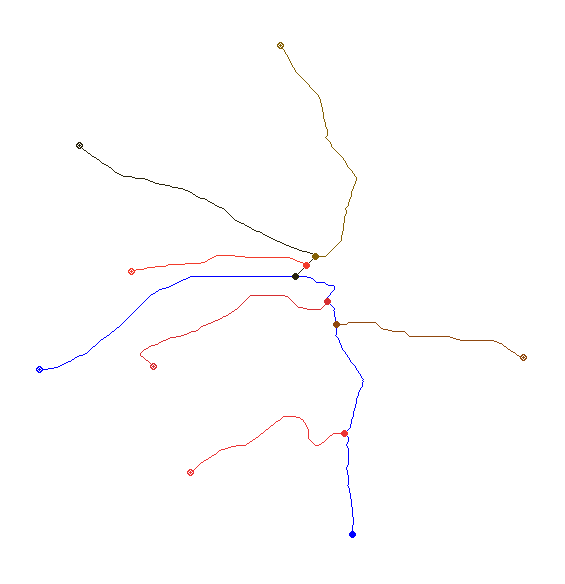}}
	\subfigure[\# 12]
	{\includegraphics[width=2.3cm]{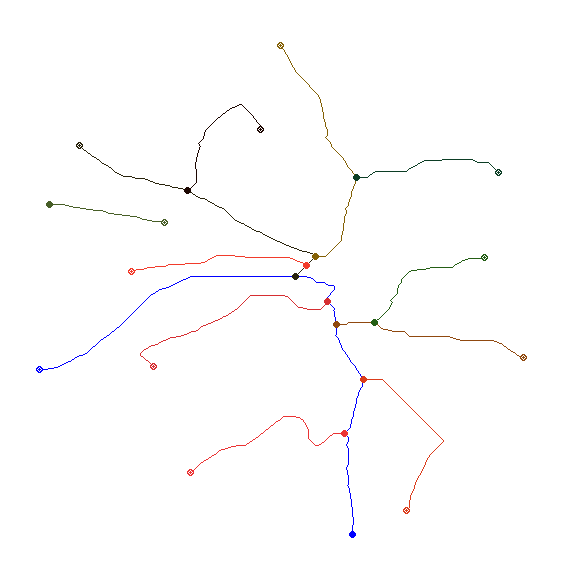}}
	\subfigure[\# 20]
	{\includegraphics[width=2.3cm]{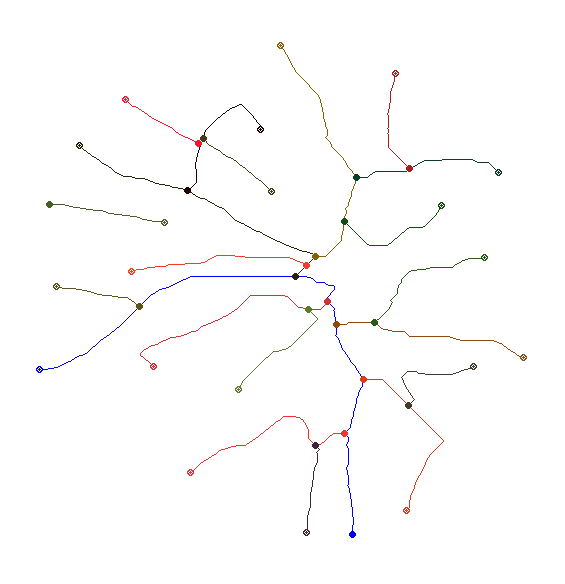}} 
	\caption {
		Intermediate steps of the curvilinear structure reconstruction for a retina image. We iteratively reconstruct the curvilinear structure according to topological importance orders. As the iteration goes on, detail structures (layer) appear.
	}\label{fig:process_example}    
\end{figure*}

\begin{figure*}[t]
	\centering	
	\subfigure{\includegraphics[width=2.3cm]{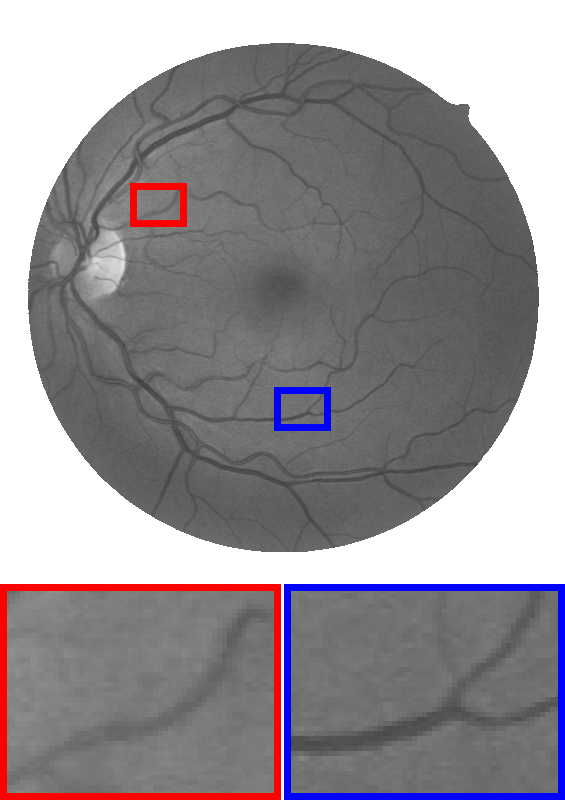}}
	\subfigure{\includegraphics[width=2.3cm]{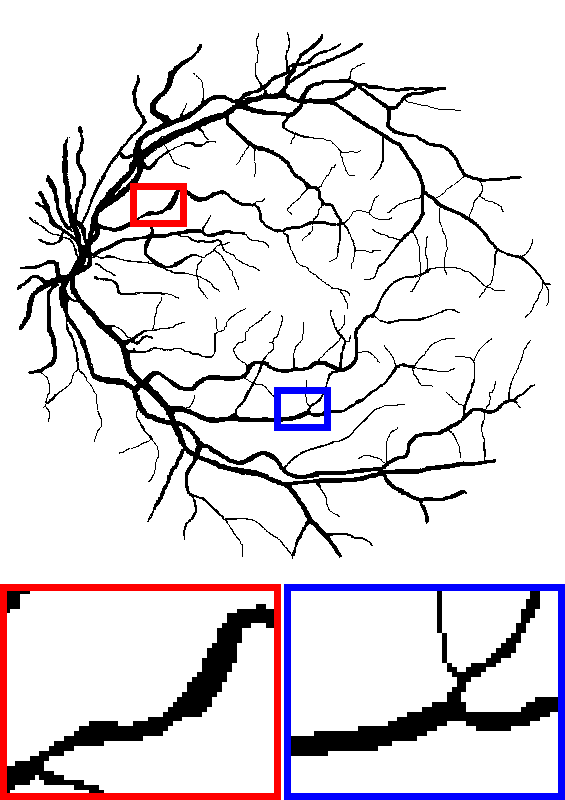}}    
	\subfigure{\includegraphics[width=2.3cm]{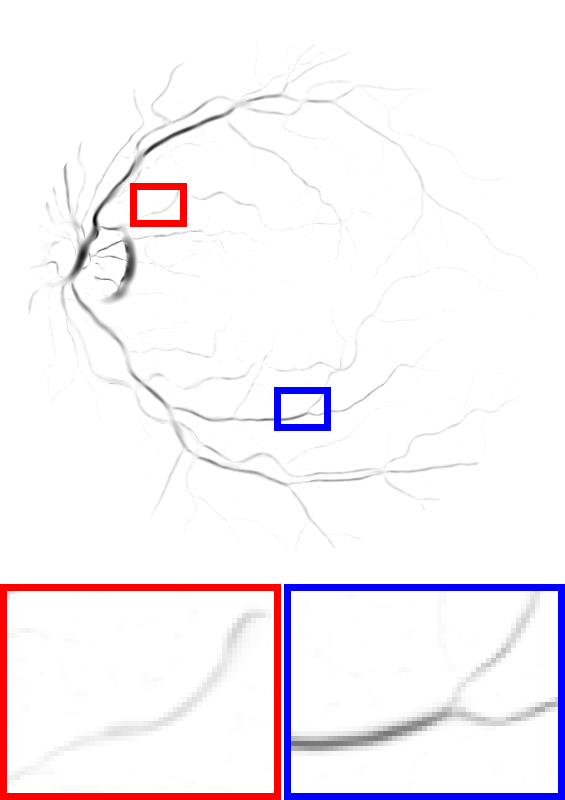}} 	
	\subfigure{\includegraphics[width=2.3cm]{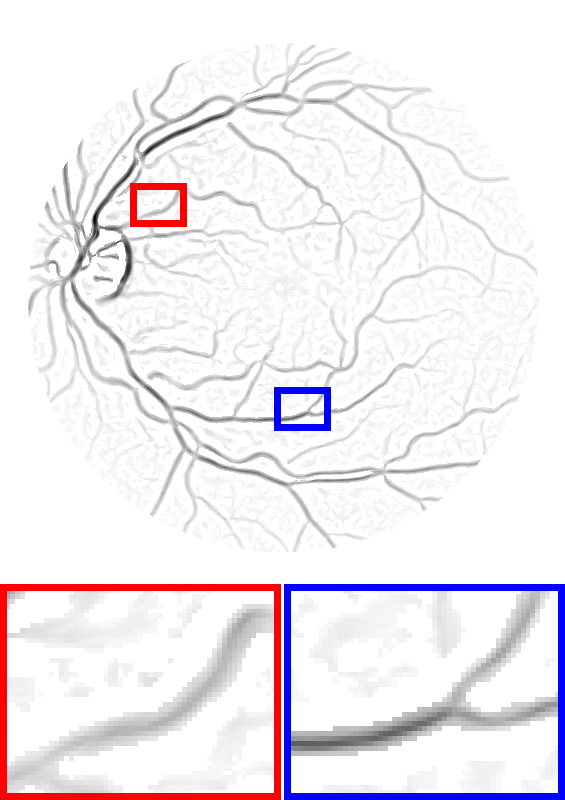}}
	\subfigure{\includegraphics[width=2.3cm]{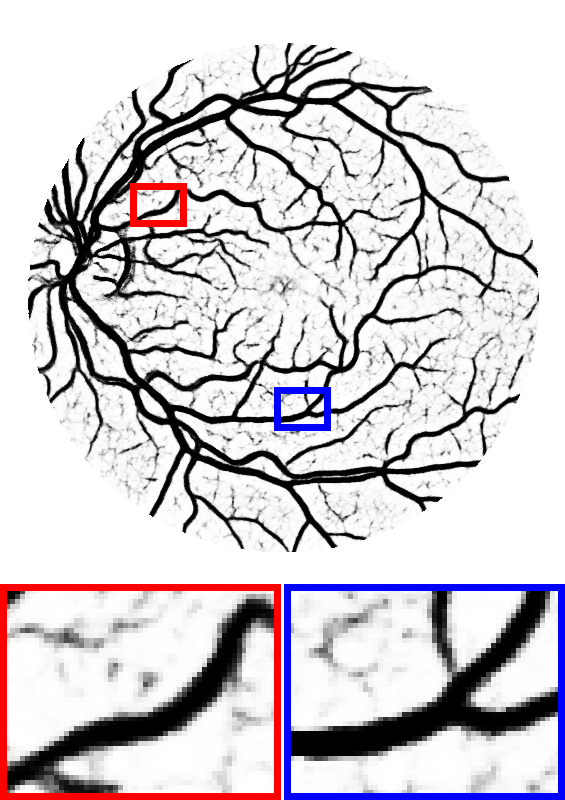}}
	\subfigure{\includegraphics[width=2.3cm]{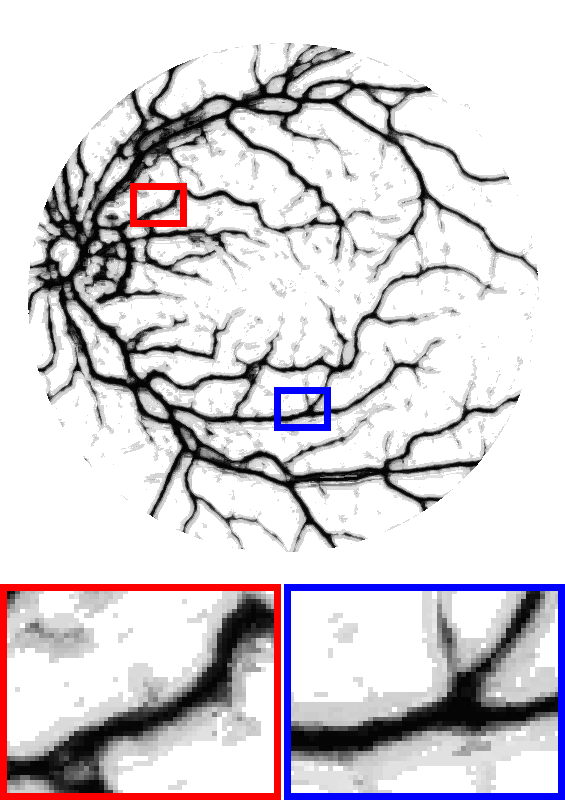}} 	
	\subfigure{\includegraphics[width=2.3cm]{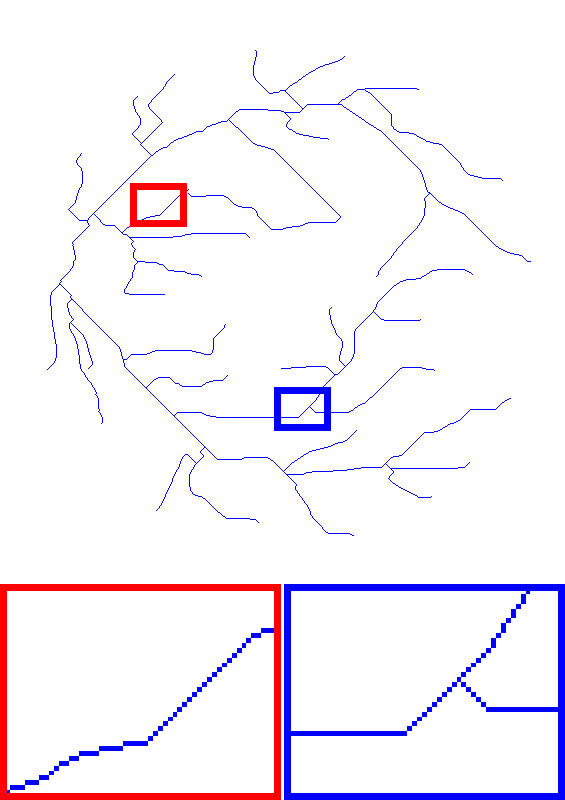}}			

	\subfigure{\includegraphics[width=2.3cm]{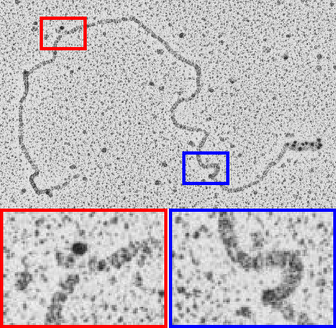}}
	\subfigure{\includegraphics[width=2.3cm]{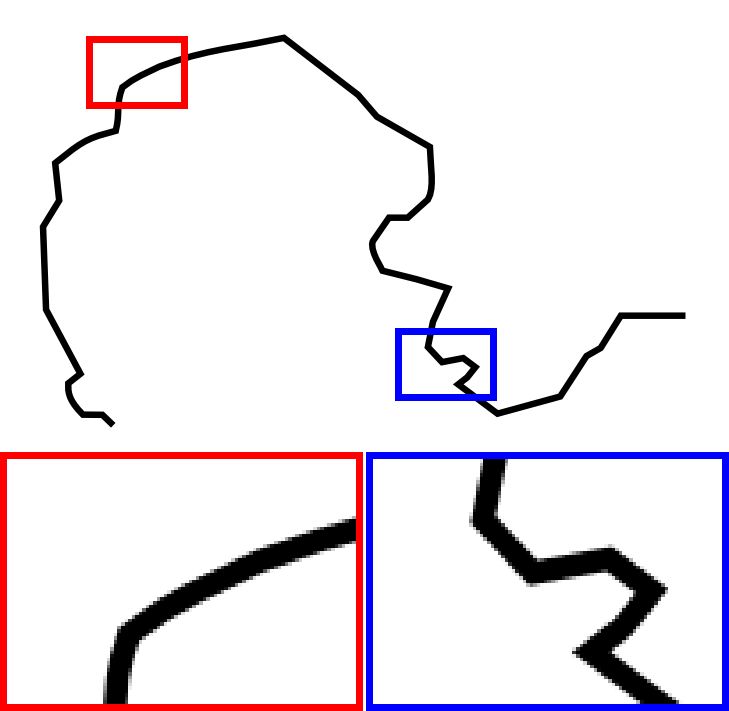}}    
	\subfigure{\includegraphics[width=2.3cm]{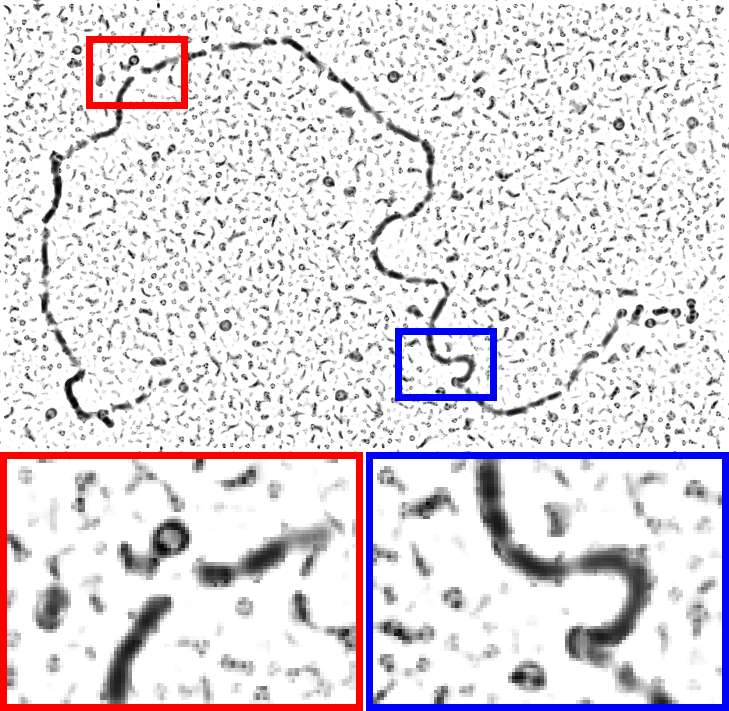}} 
	\subfigure{\includegraphics[width=2.3cm]{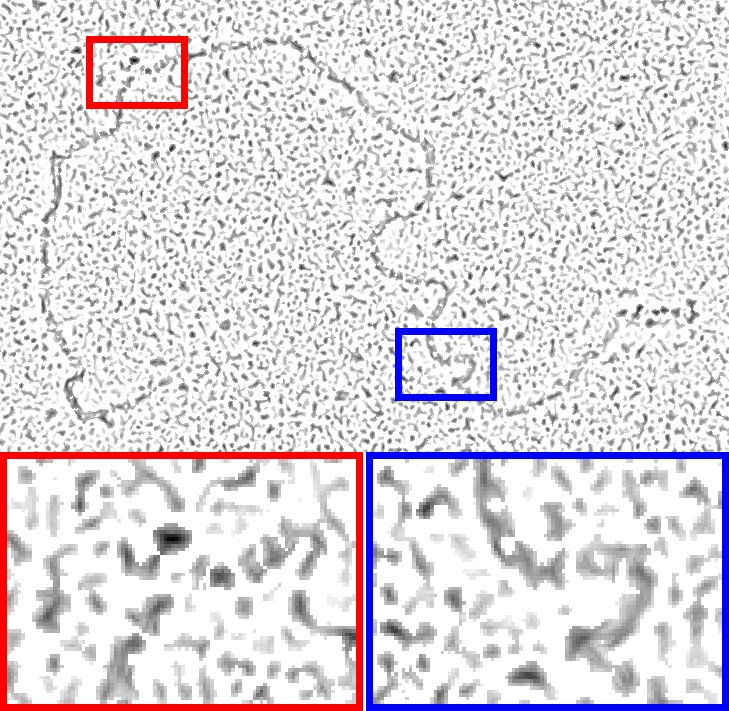}}	
	\subfigure{\includegraphics[width=2.3cm]{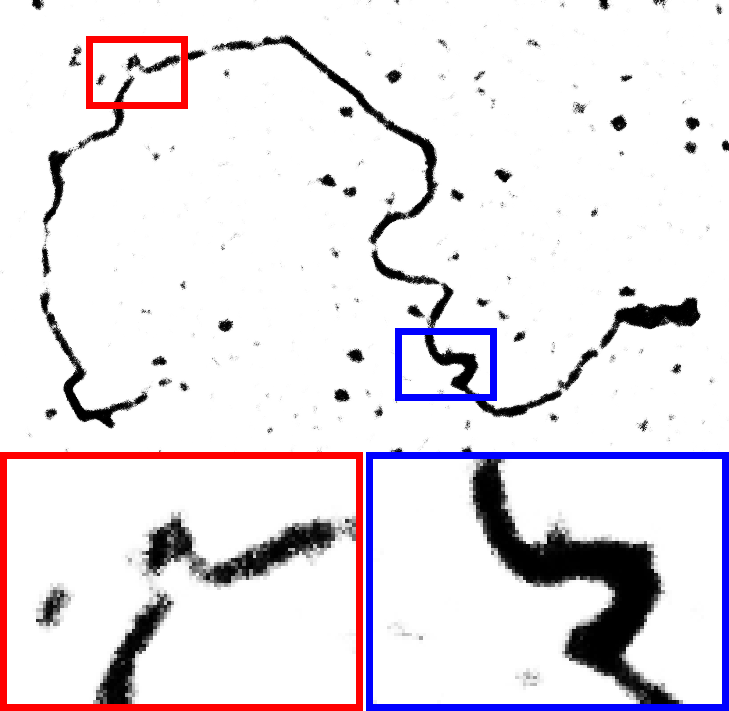}} 
	\subfigure{\includegraphics[width=2.3cm]{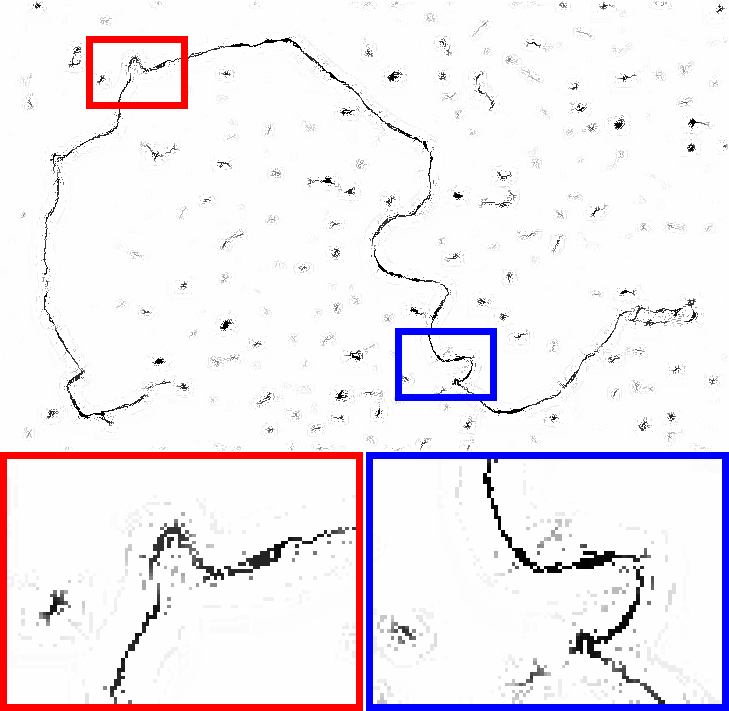}}
	\subfigure{\includegraphics[width=2.3cm]{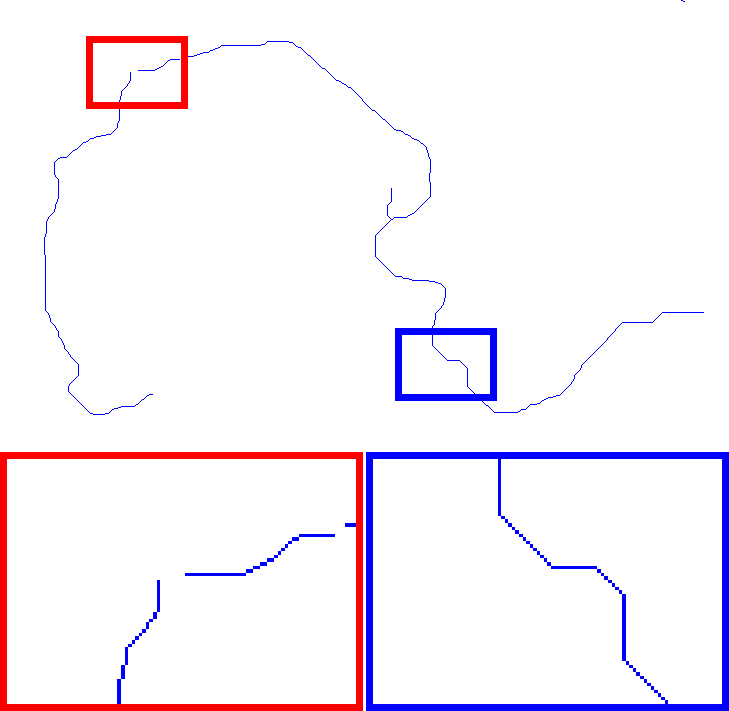}}

	\subfigure{\includegraphics[width=2.3cm]{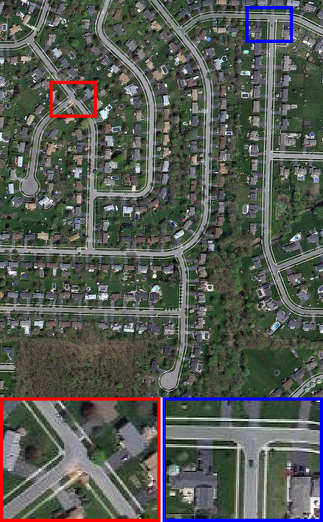}}
	\subfigure{\includegraphics[width=2.3cm]{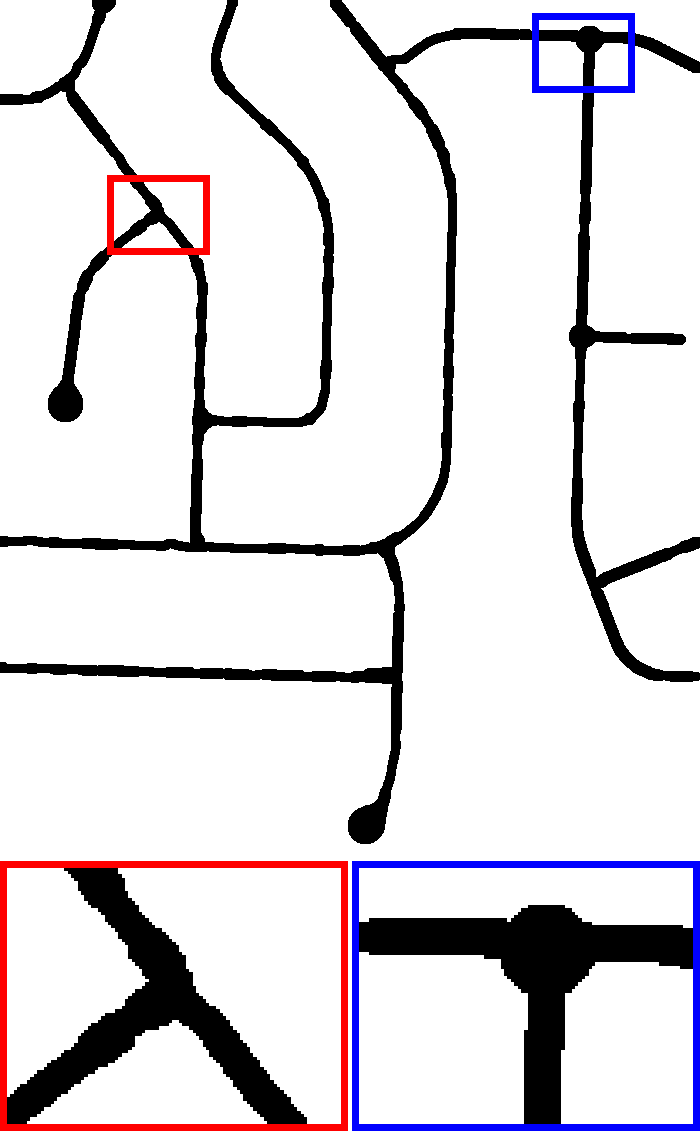}}    
	\subfigure{\includegraphics[width=2.3cm]{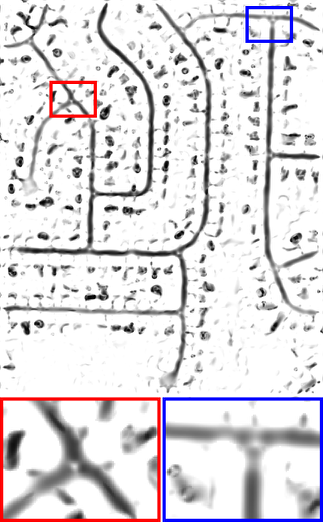}}    	
	\subfigure{\includegraphics[width=2.3cm]{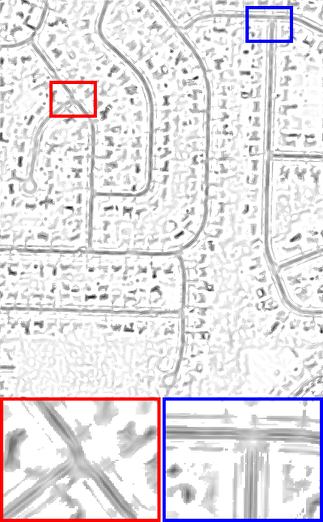}}		
	\subfigure{\includegraphics[width=2.3cm]{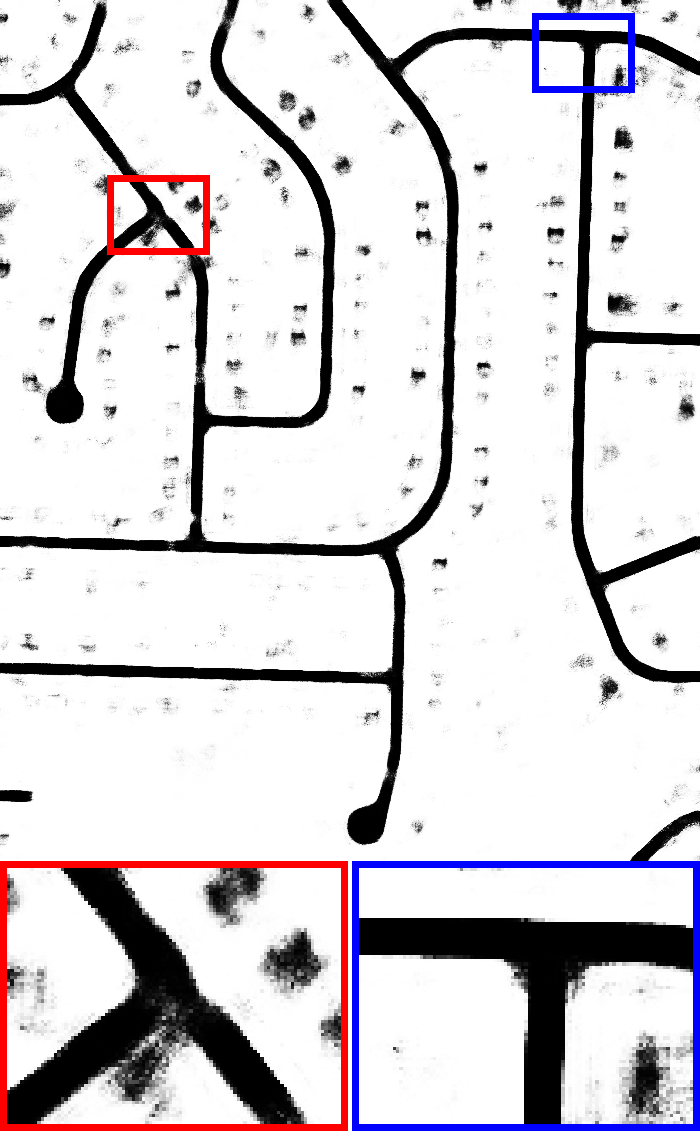}}		
	\subfigure{\includegraphics[width=2.3cm]{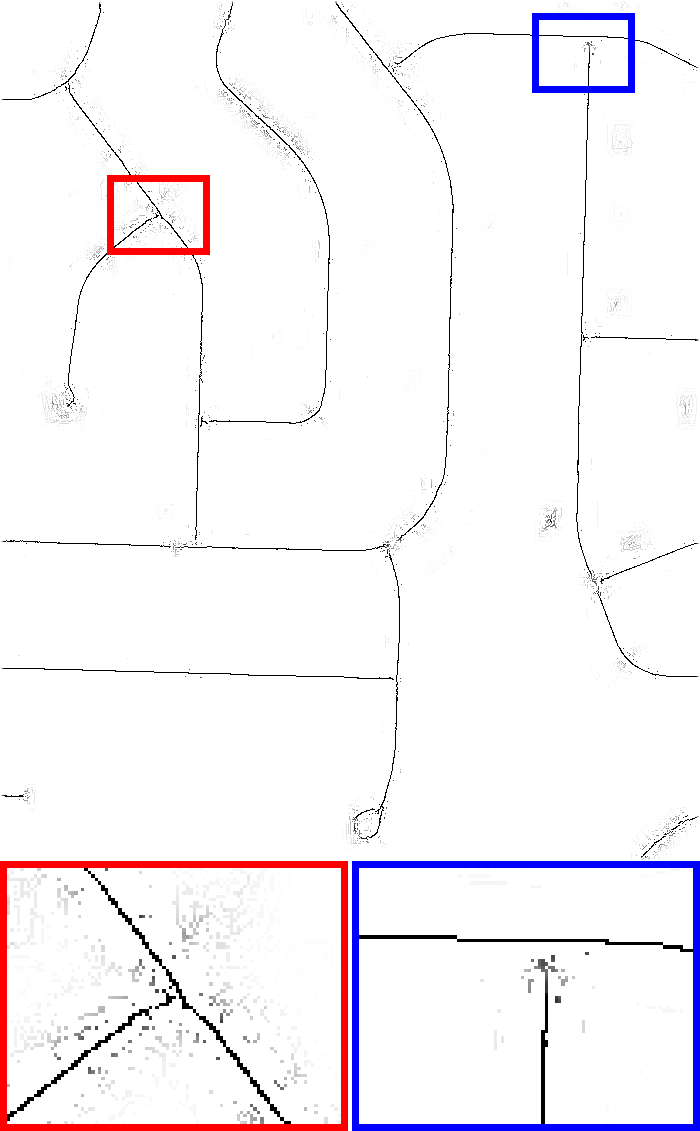}}    
	\subfigure{\includegraphics[width=2.3cm]{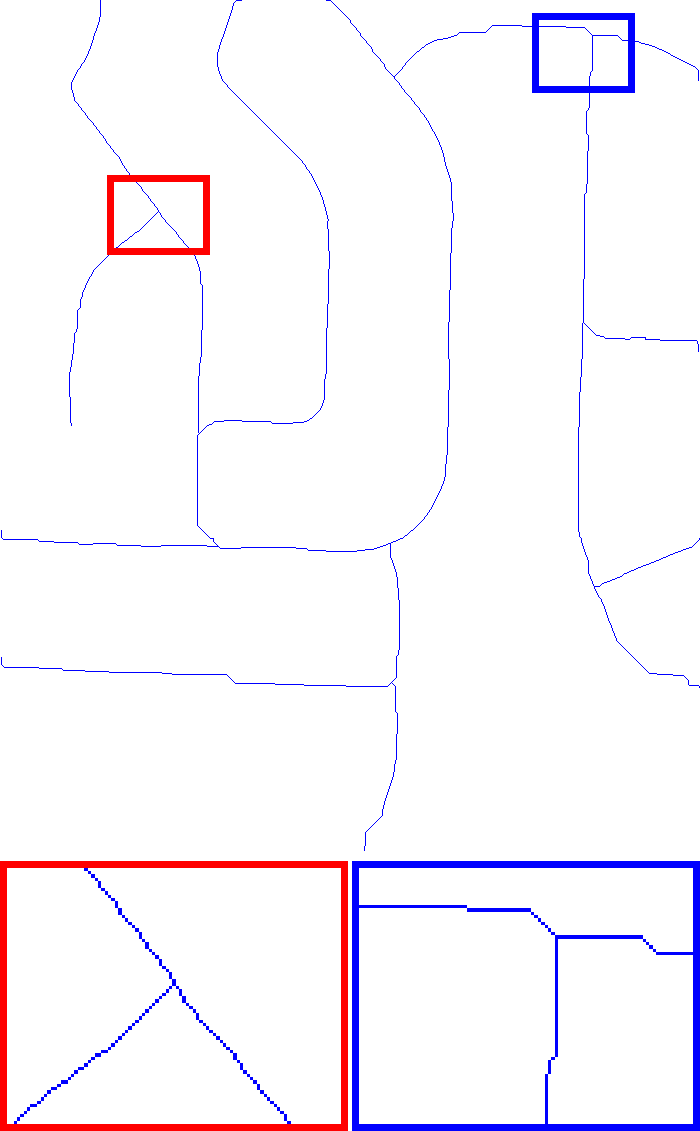}}	

	\setcounter{subfigure}{0}
	\subfigure[Input]{\includegraphics[width=2.3cm]{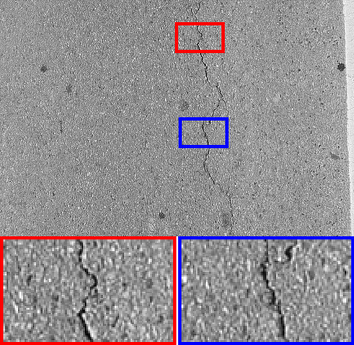}}
	\subfigure[GT]{\includegraphics[width=2.3cm]{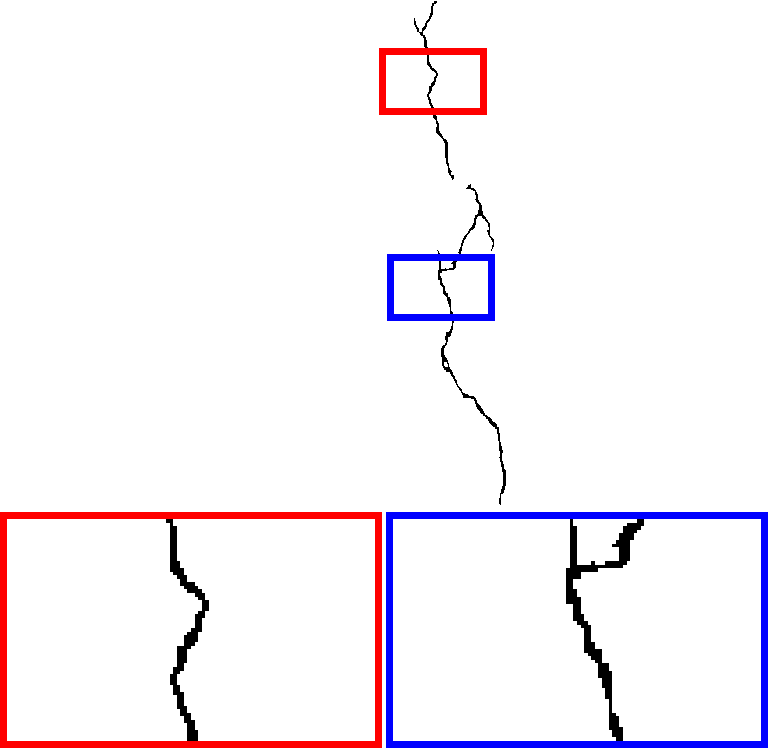}} 
	\subfigure[\cite{Frangi1998}]{\includegraphics[width=2.3cm]{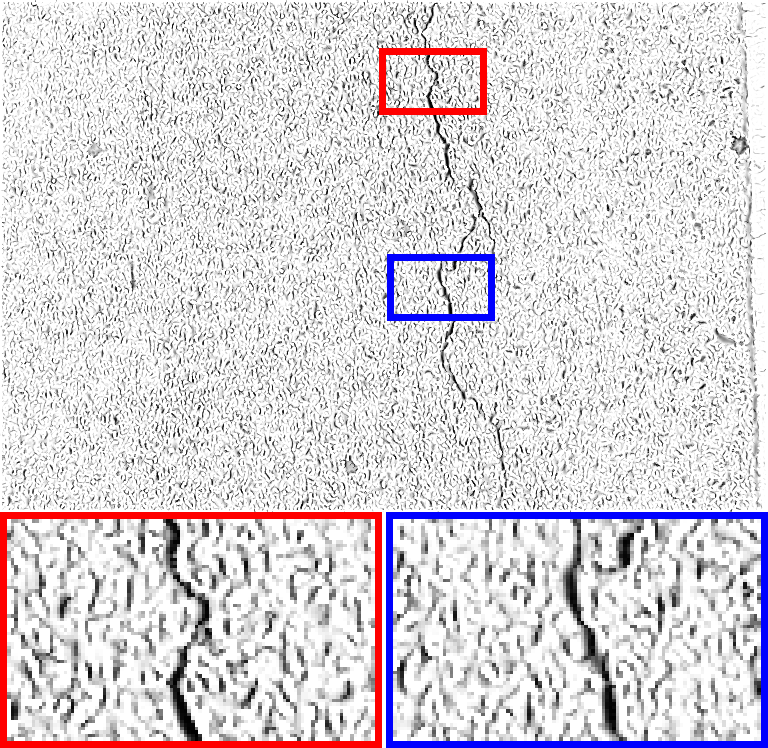}} 
	\subfigure[\cite{Law2008}]{\includegraphics[width=2.3cm]{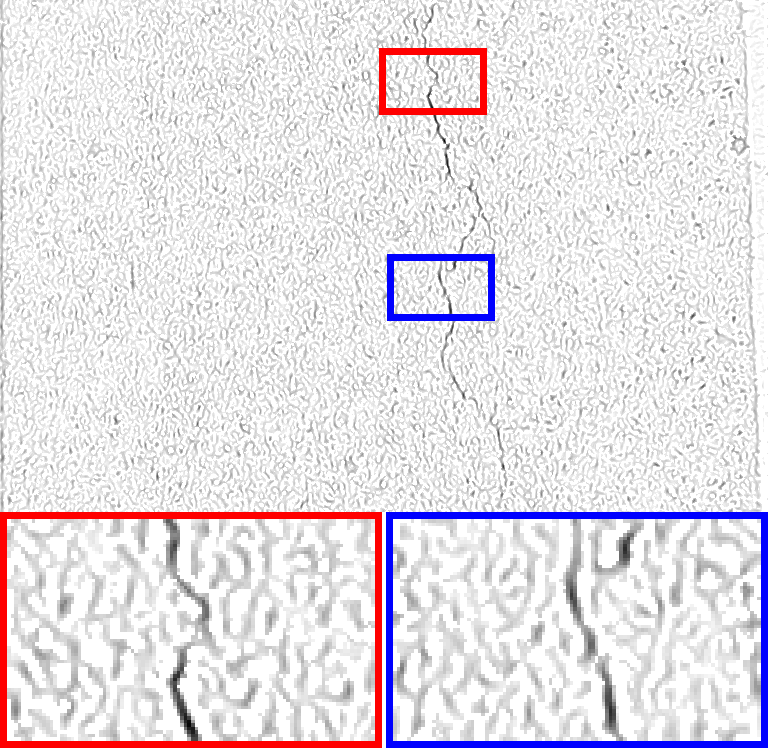}} 	
	\subfigure[\cite{Becker2013}]{\includegraphics[width=2.3cm]{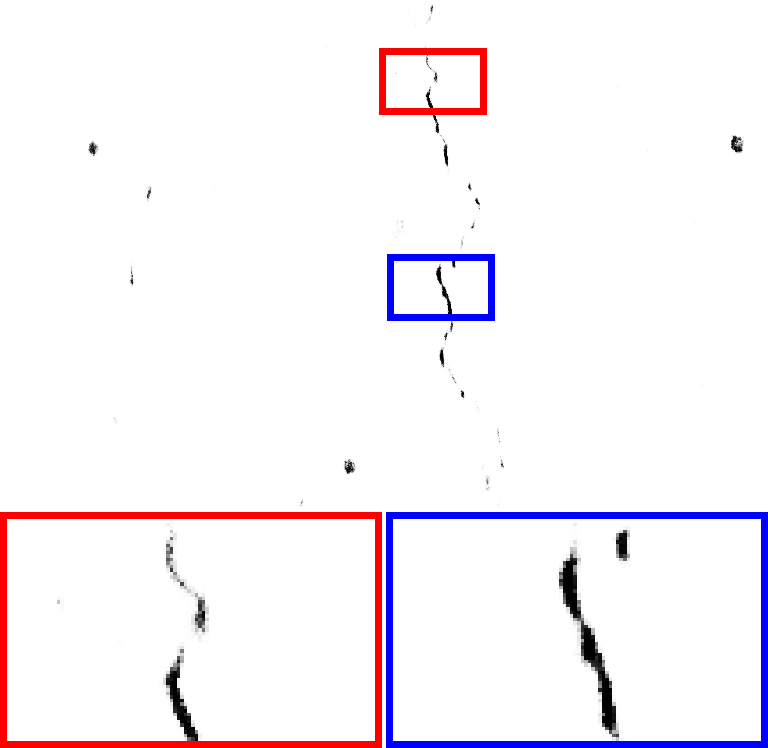}} 
	\subfigure[\cite{Sironi2014}]{\includegraphics[width=2.3cm]{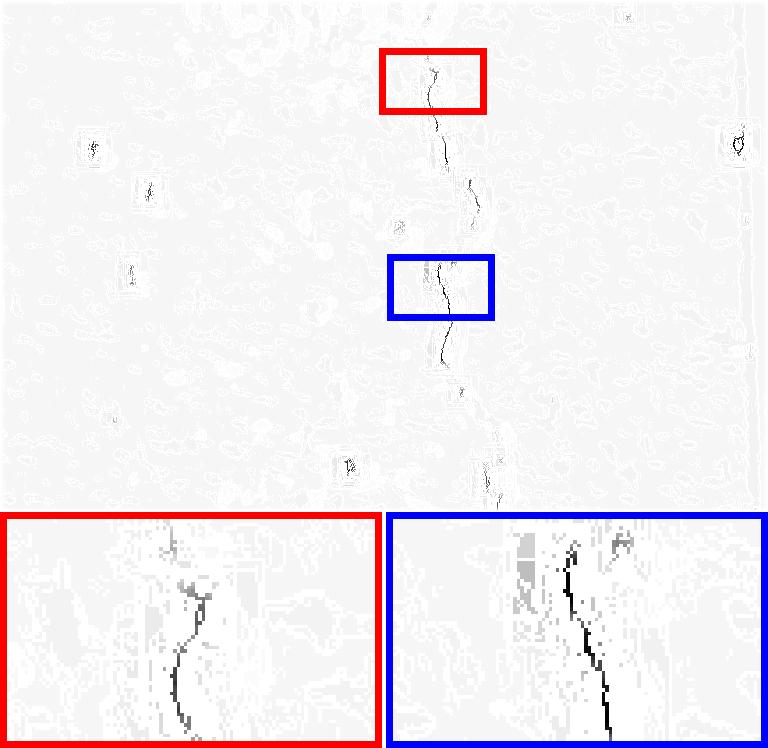}} 	
	\subfigure[Ours]{\includegraphics[width=2.3cm]{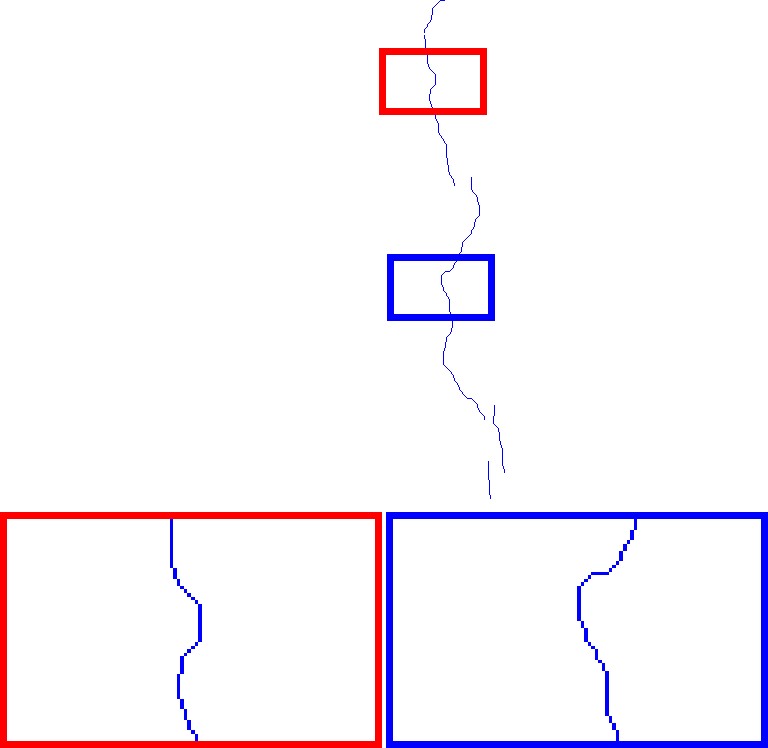}}      		
	\caption{Illustration of the curvilinear structure reconstruction results on {DRIVE}, {RecA}, {Road}, and {Crack} dataset (top to bottom). We compare the results of Frangi~\etal 's~\cite{Frangi1998}, Law $\&$ Chung's~\cite{Law2008}, Becker~\etal 's~\cite{Becker2013}, Sironi~\etal 's ~\cite{Sironi2014}, and the proposed algorithm (left to right).\label{fig:results}}
\end{figure*}

	The proposed algorithm progressively reconstructs the curvilinear  structure by adding a long path on the subgraph. Figure~\ref{fig:process_example} shows the intermediate steps of the proposed curvilinear structure simplification algorithm for {DRIVE} dataset. Unlike the previous models, the proposed algorithm is able to show different levels of detail for the latent curvilinear structure. Such information to visualize shape complexity of the curvilinear structure cannot be retrieved by setting a threshold. In practice, a few number of iterations is required to converge the algorithm and each step to find a long path takes less than milliseconds for the computation. For the experiments, we use a PC with a 2.9 GHz CPU (4 cores) and 8 GB RAM. Moreover, we visually compare the performance of the proposed algorithm with the competing algorithms. Figure~\ref{fig:results} shows the results from DRIVE, RecA, Aerial, and Cracks dataset, respectively. The proposed algorithm is the most suitable to show the topological information of the latent curvilinear structures.

	For the quantitative evaluation, we provide $F_1$ scores of the proposed algorithm and the state-of-the-art models in Table.~\ref{t:table}. The measure of true positive is sensitive for the misalignment; therefore, we consider surrounding pixels of the detection results as the true positive if a predicted point is falling into the ground truth with a small radius (equivalent to its thickness parameter $\tau$) similarly to~\cite{Sironi2014}. We also provide the average proportion of pixels to represent the curvilinear structures. It shows that the proposed algorithm efficiently draws the curvilinear structures using smaller number of pixels than the other algorithms. 
		
	The proposed algorithm achieved the best $F_1$ scores for all datasets except the {DIRVE} dataset. It is because the proposed algorithm use the fixed thickness $\tau$ to describe linear structure in the binary mask $\bf B$; however, the images consisting of {DRIVE} dataset exhibit varied thicknesses of blood vessels and many junction points. In other words, we obtain a model parameter regarding to the manually designed binary pattern $\bf B$. To overcome this drawback, we plan to exploit generative binary patterns based on the training images in the future, which is possible in our framework.

\begin{table}
    \centering
\scalebox{0.8}{    
    \begin{tabular}{l | l ||c c c c|}
\cline{2-6}
& & DRIVE & RecA & Aerial & Cracks
\\\cline{2-6}

&{Frangi~\etal~\cite{Frangi1998}} & 0.33 & 0.33 & 0.32 & 0.056\\
&{Law \& Chung~\cite{Law2008}} & 0.43 & 0.21 & 0.25 & 0.085\\
(a) $F_1$&{Becker~\etal~\cite{Becker2013}} & 0.50 & 0.45 & 0.53 & 0.23\\
&{Sironi~\etal~\cite{Sironi2014}} & \bf 0.55 & 0.50 & 0.55 & 0.27\\
&{Proposed (Graph)} & 0.36 & \bf 0.59 & \bf 0.59 & \bf 0.38\\
\cline{2-6}

&{Frangi~\etal~\cite{Frangi1998}} & 37.09 & 16.14 & 22.19 & 28.64\\
&{Law \& Chung~\cite{Law2008}} & 19.60 & 29.82 & 29.51 & 33.00\\
(b) $\rho (\%)$&{Becker~\etal~\cite{Becker2013}} & 12.67 & 9.39 & 9.76 & 32.13\\
&{Sironi~\etal~\cite{Sironi2014}} & 5.41 & 5.47 & 0.83 & 17.12\\
&{Proposed (Graph)} & \bf 0.17 & \bf 0.57 & \bf 0.35 & \bf 1.07\\\cline{2-6}
    \end{tabular}
    }\vspace{0.15cm}  
    \caption{\label{t:table}
    Comparison of the quantitative performances over the datasets. We provide results with various performance measures: (a) $F_1$ scores, and (b) average proportion $\rho$ of the pixels being a part of curvilinear structure. Boldfaced numbers are used to show the best score in each test.}    
\end{table}

	The topology of the latent curvilinear structure is varied while we compare it within the same dataset. Especially, Cracks dataset consists of difficult images due to the rough surfaced background textures. The normalization operation (\ref{eq:normalization}) is necessary to remove irregular illumination factors on the datasets. Manual segmentation (ground truth) contains many errors around boundaries and minutiae components. Also, the number of training data employed in this work is relatively small due to the difficulties of making the accurate annotations. It is remarkable that the proposed algorithm shows good performance for all datasets using the imperfect training dataset. 

\section{Conclusions}
\label{sec:conclusions}
This paper proposed a curvilinear structure reconstruction algorithm based on the ranking learning system and graph theory. The output rankings of the image patches corresponded to the plausibility of the latent curvilinear structure. Using an optimal number of pixels, the proposed algorithm provided different levels of detail during reconstruction of the curvilinear structure. More precisely, we learned a ranking function based on Structured SVM with the proposed orientation-aware curvilinear feature descriptor. In this paper, we developed a novel graphical model that infers the curvilinear structure according to the topological importance. The proposed algorithm looked for remote vertices on the subgraph which is induced from the output rankings. Across the various types of datasets, our model showed good performances to reconstruct the latent curvilinear structure with a smaller number of pixels comparing to the state-of-the-art algorithms. 


{\small
\bibliographystyle{ieee}
\bibliography{ref}

\begin{thebibliography}{10}\itemsep=-1pt

\bibitem{Arbelaez2011}
P.~Arbe\'{a}ez, M.~Maire, C.~Fowlkes, and J.~Malik.
\newblock Contour detection and hierarchical image segmentation.
\newblock {\em IEEE TPAMI}, 33(5):898--916, 2011.

\bibitem{Becker2013}
C.~Becker, R.~Rigamonti, V.~Lepetit, and P.~Fua.
\newblock Supervised feature learning for curvilinear structure segmentation.
\newblock In {\em MICCAI}, pages 526--533, 2013.

\bibitem{Bertelli2011}
L.~Bertelli, T.~Yu, D.~Vu, and B.~Gokturk.
\newblock Kernelized structural {SVM} learning for supervised object
  segmentation.
\newblock In {\em CVPR}, pages 2153--2160, 2011.

\bibitem{Borassi2015}
M.~Borassi, P.~Crescenzi, M.~Habib, W.~A. Kosters, A.~Marino, and F.~W. Takes.
\newblock Fast diameter and radius {BFS}-based computation in (weakly
  connected) real-world graphs: With an application to the six degrees of
  separation games.
\newblock {\em Theor. Comput. Sci.}, 586:59--80, 2015.

\bibitem{Chambon2010}
S.~Chambon, C.~Gourraud, J.-M. Moliard, and P.~Nicolle.
\newblock Road crack extraction with adapted filtering and {M}arkov model-based
  segmentation.
\newblock In {\em VISAPP(2)}, pages 81--90, 2010.

\bibitem{Corneil2001}
D.~G. Corneil, F.~F. Dragan, M.~Habib, and C.~Paul.
\newblock Diameter determination on restricted graph families.
\newblock {\em Discrete Applied Mathematics}, 113(2--3):143--166, 2001.

\bibitem{Frangi1998}
A.~F. Frangi, W.~J. Niessen, K.~L. Vincken, and M.~A. Viergever.
\newblock Multiscale vessel enhancement filtering.
\newblock In {\em MICCAI}, pages 130--137, 1998.

\bibitem{Freeman1991}
W.~T. Freeman and E.~H. Adelson.
\newblock The design and use of steerable filters.
\newblock {\em IEEE TPAMI}, 13(9):891--906, 1991.

\bibitem{Gonzalez2010}
G.~Gonz\'{a}lez, E.~T\"{u}retken, F.~Fleuret, and P.~Fua.
\newblock Delineating trees in noisy 2{D} images and 3{D} image-stacks.
\newblock In {\em CVPR}, pages 2799--2806, 2010.

\bibitem{Green1995}
P.~J. Green.
\newblock Reversible jump {M}arkov chain {M}onte {C}arlo computation and
  {B}ayesian model determination.
\newblock {\em Biometrika}, 82(4):771--732, 1995.

\bibitem{Hu2007}
J.~Hu, A.~Razdan, J.~C. Femiani, M.~Cui, and P.~Wonka.
\newblock Road network extraction and intersection detection from aerial images
  by tracking road footprints.
\newblock {\em IEEE TGRS}, 45(12):4144--4157, 2007.

\bibitem{Iyer2005}
S.~Iyer and S.~Sinha.
\newblock A robust approach for automatic detection and segmentation of cracks
  in underground pipeline images.
\newblock {\em Image and Vision Computing}, 23(10):921--933, 2005.

\bibitem{Jacob2004}
M.~Jacob and M.~Unser.
\newblock Design of steerable filters for feature detection using {C}anny-like
  criteria.
\newblock {\em IEEE TPAMI}, 26(8):1007--1019, 2004.

\bibitem{Jeong2015}
S.-G. Jeong, Y.~Tarabalka, and J.~Zerubia.
\newblock Marked point process model for curvilinear structures extraction.
\newblock In {\em EMMCVPR 2015, LNCS 8932}, pages 436--449, 2015.

\bibitem{Joachims2006}
T.~Joachims.
\newblock Training linear {SVM}s in linear time.
\newblock In {\em KDD}, pages 217--226, 2006.

\bibitem{Kim2014}
S.~Kim, C.~D. Yoo, S.~Nowozin, and P.~Kohli.
\newblock Image segmentation using higher-order correlation clustering.
\newblock {\em IEEE TPAMI}, 36(9):1761--1774, 2014.

\bibitem{Kohavi1995}
R.~Kohavi.
\newblock A study of cross-validation and bootstrap for accuracy estimation and
  model selection.
\newblock In {\em IJCAI}, pages 1137--1143, 1995.

\bibitem{Kwatra2005}
V.~Kwatra, I.~Essa, A.~Bobick, and N.~Kwatra.
\newblock Texture optimization for example-based synthesis.
\newblock In {\em SIGGRAPH}, pages 795--802, 2005.

\bibitem{Lacoste2005}
C.~Lacoste, X.~Descombes, and J.~Zerubia.
\newblock Point processes for unsupervised line network extraction in remote
  sensing.
\newblock {\em IEEE TPAMI}, 27(10):1568--1579, 2005.

\bibitem{Law2008}
M.~W. Law and A.~Chung.
\newblock Three dimensional curvilinear structure detection using optimally
  oriented flux.
\newblock In {\em ECCV}, pages 368--382, 2008.

\bibitem{Lucchi2013}
A.~Lucchi, Y.~Li, and P.~Fua.
\newblock Learning for structured prediction using approximate subgradient
  descent with working sets.
\newblock In {\em CVPR}, pages 1987--1994, 2013.

\bibitem{Martin2004}
D.~R. Martin, C.~C. Fowlkes, and J.~Malik.
\newblock Learning to detect natural image boundaries using local brightness,
  color, and texture cues.
\newblock {\em IEEE TPAMI}, 26(5):530--549, 2004.

\bibitem{Mittal2012}
A.~Mittal, M.~B. Blaschko, A.~Zisserman, and P.~H.~S. Torr.
\newblock Taxonomic multi-class prediction and person layout using efficient
  structured ranking.
\newblock In {\em ECCV}, pages 245--258, 2012.

\bibitem{Peng2011}
H.~Peng, F.~Long, and G.~Myers.
\newblock Automatic 3{D} neuron tracing using all-path pruning.
\newblock {\em Bioinformatics}, 27(13):239--247, 2011.

\bibitem{Perona1995}
P.~Perona.
\newblock Deformable kernels for early vision.
\newblock {\em IEEE TPAMI}, 17(5):488--499, 1995.

\bibitem{Sironi2014}
A.~Sironi, V.~Lepetit, and P.~Fua.
\newblock Multiscale centerline detection by learning a scale-space distance
  transform.
\newblock In {\em CVPR}, pages 2697--2704, 2014.

\bibitem{Staal2004}
J.~J. Staal, M.~D. Abramoff, M.~Niemeijer, M.~A. Viergever, and B.~van
  Ginneken.
\newblock Ridge based vessel segmentation in color images of the retina.
\newblock {\em IEEE TMI}, 23(4):501--509, 2004.

\bibitem{Szummer2008}
M.~Szummer, P.~Kohli, and D.~Hoiem.
\newblock Learning {CRF} using graph cuts.
\newblock In {\em ECCV}, pages 582--592, 2008.

\bibitem{Tsochantaridis2005}
I.~Tsochantaridis, T.~Joachims, T.~Hofmann, and Y.~Altun.
\newblock Large margin methods for structured and interdependent output
  variables.
\newblock {\em Journal of Machine Learning Research (JMLR)}, 6:1453--1484,
  2005.

\bibitem{Turetken2013}
E.~T\"{u}retken, F.~Benmansour, B.~Andres, H.~Pfister, and P.~Fua.
\newblock Reconstructing loopy curvilinear structures using integer
  programming.
\newblock In {\em CVPR}, pages 1822--1829, 2013.

\bibitem{Turetken2011}
E.~T\"{u}retken, G.~Gonz\'{a}lez, C.~Blum, and P.~Fua.
\newblock Automated reconstruction of dendritic and axonal tress by global
  optimization with geometric priors.
\newblock {\em Neuroinformatics}, 9(2--3):279--302, 2011.

\bibitem{Valero2010}
S.~Valero, J.~Chanussot, J.~A. Bendiktsson, H.~Talbot, and B.~Waske.
\newblock Advanced directional mathematical morphology for the detection of the
  road network in very high resolution remote sensing images.
\newblock {\em Pattern Recognition Lett.}, 31(10):1120--1127, 2010.

\bibitem{Wang2011}
Y.~Wang, A.~Narayanaswamy, and B.~Roysam.
\newblock Novel 4{D} open-curve active contour and curve completion approach
  for automated tree structure extraction.
\newblock In {\em CVPR}, pages 1105--1112, 2011.

\bibitem{Zhao2011}
T.~Zhao, J.~Xie, F.~Amat, N.~Clack, P.~Ahammad, H.~Peng, F.~Long, and E.~Myers.
\newblock Automated reconstruction of neuronal morphology based on local
  geometrical and global structural models.
\newblock {\em Neuroinformatics}, 9(2--3):247--261, 2011.

\end{thebibliography}
}

\end{document}